\documentclass{article} 
\usepackage{iclr2027_conference,times}

\usepackage{amsmath,amsfonts,bm}

\def\eqref#1{equation~\ref{#1}}

\def\1{\bm{1}}

\DeclareMathAlphabet{\mathsfit}{\encodingdefault}{\sfdefault}{m}{sl}
\SetMathAlphabet{\mathsfit}{bold}{\encodingdefault}{\sfdefault}{bx}{n}

\usepackage{hyperref}
\usepackage{url}

\usepackage{mathtools}      
\usepackage{amssymb}        
\usepackage{amsfonts}       
\usepackage{amsthm}         
\usepackage{graphicx}       
\usepackage{tikz}           
\usepackage{pifont}         
\usepackage{booktabs}       
\usepackage{array}          
\usepackage{tabularx}       
\usepackage{multirow}       
\usepackage{threeparttable} 

\usepackage{tabularx}

\usepackage{wrapfig,booktabs}

\usepackage[table]{xcolor} 

\definecolor{darker}{rgb}{0.15,0.45,0.4} 
\definecolor{green}{HTML}{009B55}       
\definecolor{nblue}{cmyk}{0.95,0.0,0.2,0.2} 
\definecolor{projectlink}{HTML}{3588AC}
\newcommand{\projectglobe}{%
  \tikz[baseline=-0.35em,x=1.05em,y=1.05em]{%
    \draw[projectlink,line width=0.55pt] (0,0) circle (0.5);
    \draw[projectlink,line width=0.45pt] (0,0) ellipse (0.22 and 0.5);
    \draw[projectlink,line width=0.45pt] (-0.5,0)--(0.5,0);
    \draw[projectlink,line width=0.45pt] (-0.43,0.25)--(0.43,0.25);
    \draw[projectlink,line width=0.45pt] (-0.43,-0.25)--(0.43,-0.25);%
  }%
}

\definecolor{codegreen}{rgb}{0,0.6,0}
\definecolor{codegray}{rgb}{0.5,0.5,0.5}
\definecolor{codepurple}{rgb}{0.58,0,0.82}
\definecolor{backcolour}{rgb}{0.98,0.98,0.97}

\hypersetup{
  colorlinks=true,
  urlcolor=darker,
  citecolor=darker,
  linkcolor=darker
}

\usepackage{cleveref}

\usepackage{enumitem}  
\usepackage{microtype} 
\usepackage{xspace}    
\usepackage{upquote}   

\usepackage{listings}  
\usepackage{fancyvrb}  
\usepackage{tcolorbox} 
\tcbuselibrary{listings,breakable,skins} 

\RecustomVerbatimCommand{\VerbatimInput}{VerbatimInput}{%
  fontsize=\footnotesize,
  frame=single,
  framesep=0.5em,
  labelposition=topline,
}

\lstdefinestyle{prompt}{%
  basicstyle=\ttfamily\small,
  breaklines=true,
  frame=none,
  keywordstyle=\bfseries,
  showstringspaces=false,
  literate={~} {$\sim$}{1},
}

\tcbset{paperprompt/.style={
  enhanced jigsaw,
  breakable,
  width=\linewidth,
  arc=1.5pt,
  boxrule=0.45pt,
  colback=white,
  colframe=darker!28!black!35,
  colbacktitle=darker!8!white,
  coltitle=black,
  coltext=black,
  fonttitle=\rmfamily\bfseries\fontsize{10}{11}\selectfont,
  fontupper=\rmfamily\fontsize{9.2}{10.6}\selectfont,
  boxsep=0pt,
  left=9pt,
  right=9pt,
  top=6pt,
  bottom=6pt,
  toptitle=4pt,
  bottomtitle=4pt,
  before skip=6pt,
  after skip=8pt,
  pad at break*=4pt,
  lines before break=4,
  before upper={\setlength{\parindent}{0pt}\setlength{\parskip}{3pt}
    \hyphenpenalty=10000\emergencystretch=1em},
}}
\newcommand{\promptheading}[1]{%
  \par\addvspace{4pt}\noindent\textbf{#1}\par\nobreak}

\theoremstyle{definition}

\theoremstyle{remark}

\title{Scaling Agentic Data for Long-Horizon Terminal Intelligence}

\newcommand{\projectauthors}{%
\normalfont
\parbox[t]{\dimexpr\textwidth-2\tabcolsep\relax}{%
  \setlength{\parskip}{0pt}
  \setlength{\parindent}{0pt}
  \normalfont\normalsize\raggedright
  {\bfseries
    \addtolength{\baselineskip}{2pt}
    \mbox{Shun Zou$^{1,2*}$}\hspace{0.7em}%
    \mbox{Ziyu Ma$^{2*}$}\hspace{0.7em}%
    \mbox{Yi Zou$^{1,2*}$}\hspace{0.7em}%
    \mbox{Yong Wang$^{2\dagger\ddagger}$}\hspace{0.7em}%
    \mbox{Lin Chen$^{1}$}\hspace{0.7em}%
    \mbox{Zehui Chen$^{1}$}\hspace{0.7em}%
    \mbox{Guanhua Chen$^{3}$}\hspace{0.7em}%
    \mbox{Xiangxiang Chu$^{2}$}\hspace{0.7em}%
    \mbox{Feng Zhao$^{1\ddagger}$}\par
  }
  \vspace{6pt}
  $^{1}$University of Science and Technology of China \\
  $^{2}$AMAP, Alibaba Group \\
  $^{3}$Southern University of Science and Technology\par
  \vspace{6pt}
  \href{https://zs1314.github.io/TerminalHorizon/}{%
    \textcolor{projectlink}{\projectglobe\hspace{0.35em}Project Page}}
  \qquad
  \href{https://huggingface.co/collections/shunzou05/terminalhorizon}{%
    \raisebox{-0.2em}{\includegraphics[height=1.15em]{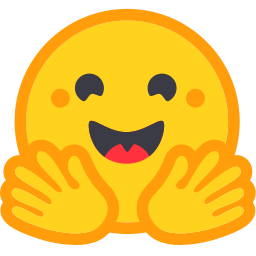}}\hspace{0.35em}%
    \textcolor{projectlink}{Data \& Model}}\par
}
}
\author{\projectauthors}

\iclrfinalcopy 
\begin{document}

\maketitle
\begingroup
\renewcommand{\thefootnote}{\fnsymbol{footnote}}
\footnotetext[1]{Equal contribution. Work done during internship at AMAP, Alibaba Group.} 
\footnotetext[2]{Project lead.\hspace{0.5em}$^\ddagger$Corresponding authors.}
\endgroup
\vspace{-12pt}

\begin{abstract}
Terminal agents are moving beyond local operations toward completing real-world professional tasks.
Short trajectories can demonstrate commands and tool calls, but they do not fully show how agents accumulate evidence through exploration, use it to inform judgments, continually adjust strategies based on feedback, and turn intermediate results into final deliverables.
High-quality long-horizon trajectories capture these \textit{agentic behavioral patterns} and their connections within a single problem-solving process, providing richer experience for sustained autonomous work.
However, environment synthesis may complete or simplify the key work underlying these patterns in advance and thus remove the need for exploration, implementation, and verification that the agent was meant to perform.
To address this, we introduce \textbf{TerminalHorizon}, a fully automated data synthesis engine for long-horizon terminal agents.
Starting from real-world professional work, it uses independent execution to identify key work omitted during construction and uses the resulting evidence to guide environment reconstruction.
To extend these patterns across stages, it progressively expands tasks under the same goal, introducing new requirements that build on prior results and call for further exploration and improvement.
Using this engine, we construct 1.5K task environments and perform supervised fine-tuning of Qwen3.5-27B on the resulting dataset to obtain \textbf{TerminalHorizon-27B}.
We achieve substantial performance gains on all 12 benchmarks spanning coding-agent tasks, agentic tasks in real-world production settings, and general reasoning.
Multiple RSI case studies, each with a 12-hour research budget, further demonstrate the model's ability to continue experimenting beyond an initial solution, improve its methods based on feedback, and retain effective improvements in its final deliverables.

\end{abstract}

\begin{figure}[!ht]
\vspace{-2ex}
    \centering
    \includegraphics[width=\linewidth]{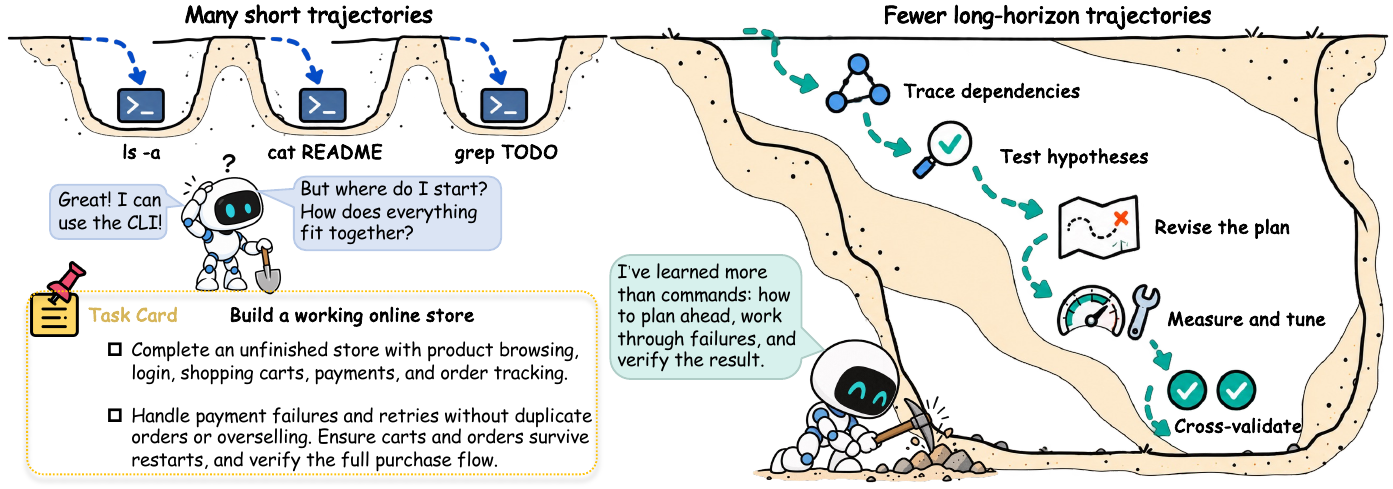}
    \vspace{-3ex}
\caption{Motivation for TerminalHorizon. Short trajectories can teach agents to perform terminal operations, but not to autonomously complete complex, long-horizon tasks. Our long-horizon trajectories connect agentic behavioral patterns, such as exploration, strategy revision, and verification, into a complete autonomous problem-solving process.
}
    \label{fig:teaser}
    \vspace{-3ex}
\end{figure}

\section{Introduction}
\label{sec:intro}
\vspace{-1ex}
LLM agents are moving beyond answering questions in dialogue toward autonomous work in digital environments~\citep{AutoLab,Multi-Agent,dong2026longhorizon,OneDayAgent}. The terminal provides a general interface for this work by bringing code, data, tools, and execution feedback into a shared workspace~\citep{Terminal-Bench}, enabling agents to tackle complex tasks in software development, system administration, and scientific computing~\citep{anthropic2025claudecodeagentic,openai2025codexcli}. As their scope expands from individual operations to long-horizon tasks~\citep{dong2026longhorizon,ma2026longhorizon,qwen38,AgentFold}, agents need to coordinate interdependent stages and adapt to execution feedback to complete these tasks reliably~\citep{SWE-Bench-Pro,Long-Horizon-Terminal-Bench,SWE-Marathon,ProgramBench}. Developing these capabilities requires rich interaction experience during training, but high-quality terminal training data remains scarce.

Recent work on terminal environment synthesis constructs executable and verifiable tasks at scale through skill composition, task generation grounded in technical evidence, and recursive expansion, allowing agents to gain interaction experience through execution and feedback~\citep{wu2026terminaluniverseturningagenttrajectories,DBLP:journals/corr/abs-2605-20876,RST,wu2026large,DBLP:journals/corr/abs-2604-25727}. These methods expand task coverage, but their trajectories remain relatively short on average as shown in Figure~\ref{fig:interaction-turns-comparison}, offering limited experience in sustained problem solving. Short trajectories can capture useful agentic behavioral patterns over a few steps, but the same patterns can extend across multiple stages or even an entire trajectory in long-horizon tasks. For example, agents may repeatedly revise their plans as evidence accumulates, build on intermediate results through a sequence of dependent steps, and revisit earlier work when later verification reveals problems that require further diagnosis and correction. This motivates our focus on high-quality long-horizon trajectories that capture these longer agentic behavioral patterns and show how they work together to complete a task (Section~\ref{agentic patterns}).

However, synthesizing these long-horizon trajectories presents two challenges.
(i) Complex tasks do not necessarily lead to the intended agentic behavioral patterns.
A task may require agents to investigate a problem, develop a solution, and verify it, yet supplied materials may already solve parts of the problem, allowing agents to skip exploration or implementation. These gaps can prevent the intended patterns from fully developing in the collected trajectories.
(ii) Even when these patterns appear, long-horizon training data should also capture how they unfold over longer problem-solving processes and connect across stages.
More difficult tasks or more interaction turns do not necessarily extend these patterns or strengthen their connections. The challenge is to identify and implement new requirements that go beyond what the original solution can achieve, while keeping the same goal. The added work needs to build on earlier judgments, actions, and results, so that these patterns continue to develop as the task progresses.

To this end, we introduce \textbf{TerminalHorizon}, a fully automated data synthesis engine with two stages: Challenge-Guided Long-Horizon Synthesis and Progressive Horizon Expansion. The first stage converts real-world professional problems into executable tasks and uses an agent uninvolved in their construction to solve them. Analysis of the resulting trajectories checks whether the intended  agentic behavioral patterns occur and identifies where task construction has simplified or already completed the analysis, implementation, or verification expected of the agent. These findings guide environment reconstruction so that the agent must perform this work during execution. The second stage expands the original task under the same goal, introducing requirements that call for further analysis, implementation, and verification. It makes this added work depend on earlier results and execution feedback, allowing the intended patterns to extend across stages. Together, these stages support the generation of long-horizon trajectories in which the intended agentic behavioral patterns emerge and extend across stages toward task completion.


Using TerminalHorizon, we construct 1,535 task environments and collect 2,852 training trajectories to form \textbf{TerminalHorizon-3K}. The trajectories average 52.8 interaction turns, substantially longer than those from the methods compared in Figure~\ref{fig:interaction-turns-comparison}. Supervised Fine-tuning Qwen3.5-27B on these 2,852 trajectories yields \textbf{TerminalHorizon-27B}, which improves over the base model by 14.2\% and 13.85\% on Terminal-Bench 2.0 and 2.1, respectively. These gains extend to real-world productivity tasks, with improvements of 11.7\% in Pass Rate and 23.3\% in Partial score on AutomationBench, and 11.3\% on VitaBench. Furthermore, case studies on multiple long-horizon recursive self-improvement (RSI) tasks, each with a 12-hour research budget, show how the model continues experimenting beyond an initial solution, refines its methods based on feedback, and incorporates these improvements into final deliverables. Our main contributions are as follows:
\begin{itemize}[leftmargin=*,itemsep=.5pt,topsep=0pt,parsep=0pt]
    \item We propose a perspective on high-quality long-horizon terminal data centered on agentic patterns and their connections across stages, emphasizing the substantive work that environments must preserve to elicit them.
    \item We introduce \textbf{TerminalHorizon}, a fully automated data synthesis engine that combines challenge-guided construction with progressive expansion to build \textbf{TerminalHorizon-3K}, comprising 1,535 task environments and 2,852 training trajectories.
    \item Supervised Fine-tuning on only about 3K trajectories improves performance across 12 benchmarks spanning coding, real-world productivity, and general reasoning, while multiple 12-hour RSI case studies demonstrate continued experimentation, improvement, and final delivery.
\end{itemize}

\section{Method}
\label{sec:method}

\begin{figure}[t]
    \centering
    \includegraphics[width=\linewidth]{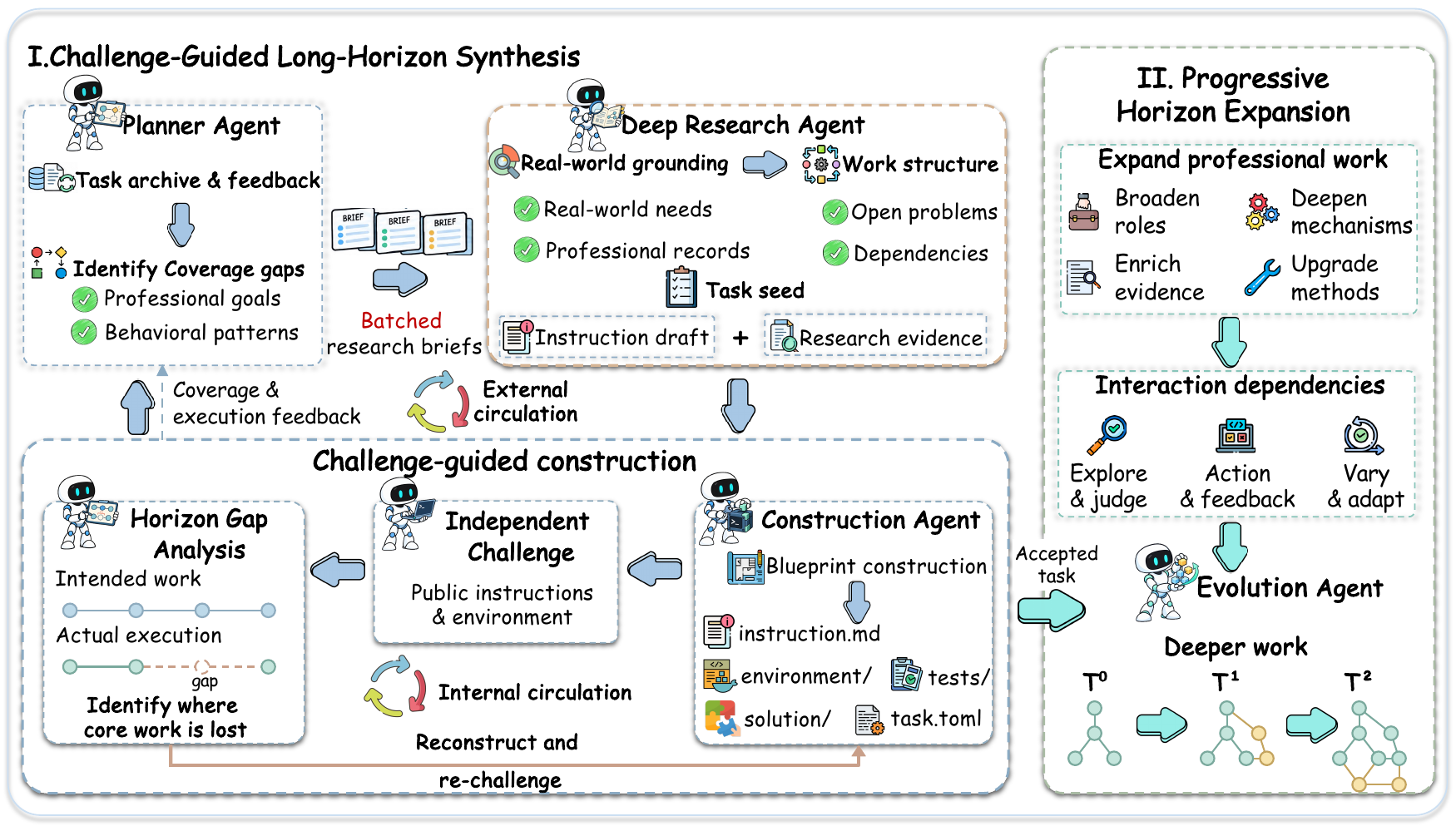}
    \vspace{-5ex}
    \caption{
    Overview of TerminalHorizon.
    Challenge-guided synthesis refines task environments through independent execution to preserve core problem-solving work.
    Progressive horizon expansion deepens tasks along the same goal to support richer agentic behavioral patterns.
    }
    \label{fig:pipline}
    \vspace{-3ex}
\end{figure}

TerminalHorizon builds executable environments from real-world professional problems to elicit agentic behavioral patterns (Section~\ref{agentic patterns}).
In the first stage (Figure~\ref{fig:pipline}), independent execution reveals work simplified or completed in construction and guides reconstruction.
The second stage extends accepted tasks under the same goal, making added work depend on prior results and feedback.

\subsection{Task Formulation and Design Objective}
\label{sec:method-overview}
\noindent\textbf{Task formulation.}
Both stages construct and refine executable terminal task packages, each comprising five components~\citep{Terminal-Bench}:
\begin{equation}
  T=(I,E,S,V,M),
  \qquad
  \mathcal P(T)=(I,E).
\end{equation}
The public instruction $I$ specifies the goal and constraints, while the executable environment $E$ provides the initial state, tools, and execution mechanisms.
The reference solution $S$, private verifier $V$, and execution configuration $M$ support task construction and evaluation.
During execution, the agent has access only to $\mathcal P(T)$ and gathers information through interactions with the environment.

\noindent\textbf{Long-horizon design objective.}
The work we aim to preserve connects exploration, diagnosis, construction, verification, and strategic coordination, rather than only demonstrating individual terminal operations.
We describe these processes using 27 agentic behavioral patterns across five families (Section~\ref{agentic patterns}).
For example, relationship tracing follows connections between objects to inform decisions.
The taxonomy is shown in Figure~\hyperref[fig:data]{\ref*{fig:data}d}, with full definitions in Appendix~\ref{app:agentic-behavioral-patterns}.
Environment design must therefore retain evidence to explore, analysis and implementation left to complete, and feedback that helps agents revise earlier judgments.

\subsection{Challenge-Guided Long-Horizon Synthesis}
\label{sec:basecamp}
An inner loop uses independent execution to guide task reconstruction, while an outer loop uses the resulting tasks and review evidence to plan subsequent research batches.

\noindent\textbf{Autonomous Batch Planning.}
Each batch begins by identifying professional work that would broaden coverage of agentic behavioral patterns.
The Planner Agent uses the goals, solution approaches, and feedback in existing tasks, together with prior research and review records, to select the next research directions.
For batch $b$, let $\mathcal D_b$ be the accepted task set, $\mathcal F_b$ the accumulated records, and $n_b$ the number of available research slots.
The Planner produces research briefs:
\begin{equation}
  \mathcal Q_b
  = \operatorname{Plan}(\mathcal D_b,\mathcal F_b;n_b),
  \qquad
  \mathcal Q_b=\{q_{b,i}\}_{i=1}^{n_b}.
  \label{eq:batch-synthesis}
\end{equation}
Each brief $q_{b,i}$ specifies a scope of work, requirements not yet covered by existing tasks, and real-world sources to investigate, providing the starting point for research.

\noindent\textbf{Evidence-Grounded Deep Research.}
A research brief sets the direction, but task construction requires a concrete account of the work involved.
Following $q_{b,i}$, the Deep Research Agent investigates documentation, code repositories, issue reports, and professional materials to establish real-world needs, technical constraints, and evidence needed to assess completion.
The research identifies what the agent needs to discover, which problems require further analysis, implementation, or debugging, and how one result supports subsequent work.
These findings form a task seed $z_{b,i}$, containing a draft public instruction and supporting evidence, which is passed to environment construction.

\noindent\textbf{Joint Environment Construction.}
The task seed specifies the intended work, which construction must support without completing it in advance.
From $z_{b,i}$, the Construction Agent synthesizes a concrete instance with a controlled initial state and verification conditions, specifying the available materials and tools, the work left to the agent, and the completion criteria.
It then jointly constructs the instruction, environment, reference solution, verifier, and execution configuration around these requirements, producing a candidate package $T$.
Evaluation must follow the public task requirements:
\begin{equation}
  \mathcal C_{\mathrm{sem}}(V)
  \subseteq \operatorname{Sem}\!\left(I,\mathcal O_{\mathrm{pub}}(E)\right).
  \label{eq:public-contract}
\end{equation}
The left-hand side denotes the completion requirements checked by the verifier, and the right-hand side those supported by $I$ and discoverable evidence $\mathcal O_{\mathrm{pub}}(E)$.
Test data may remain private, but scoring cannot introduce requirements the agent cannot discover.
The reference solution gives one feasible approach, while the verifier accepts other implementations meeting the public requirements.

\noindent\textbf{Independent Challenge.}
An executable package does not establish that the intended work remains necessary during task solving.
We therefore ask an Independent Challenger, uninvolved in construction, to attempt the candidate $T_r$ at round $r$.
Its policy $\pi_B$ accesses only $\mathcal P(T_r)$, without research records, the blueprint, the reference solution, or private tests.

At each step, policy $\pi=\pi_B$ selects an action given the public instruction and history $h_t=(o_0,a_0,\ldots,o_t)$, and the environment dynamics $P_E$ update the state and return an observation:
\begin{equation}
  a_t \sim \pi(\cdot \mid I,h_t),
  \qquad
  (s_{t+1},o_{t+1}) \sim P_E(\cdot \mid s_t,a_t).
  \label{eq:terminal-interaction}
\end{equation}
The resulting trajectory $\tau_r=(o_0,a_0,\ldots,a_{H_r-1},o_{H_r})$ records $H_r$ interactions, and the verifier evaluates the final state and artifacts to compute $\rho_r=V_r(s_{H_r})$.
The complete package, execution trace, and scoring evidence then support a diagnosis:
\begin{equation}
  \tau_r \sim \operatorname{Rollout}(\pi_B;\mathcal P(T_r)),
  \qquad
  d_r = \mathcal A_{\mathrm{syn}}(T_r,\tau_r,\rho_r).
  \label{eq:synthesis-challenge-audit}
\end{equation}
Here, $\mathcal A_{\mathrm{syn}}$ is the synthesis review, and $d_r$ records runtime defects, information leakage, scoring inconsistencies, and whether the intended work is preserved, providing evidence for reconstruction.

\noindent\textbf{Horizon Gap Analysis and Reconstruction.}
The trajectory reveals what the agent actually did, but reconstruction also requires identifying which gaps arise from task construction.
Horizon Gap Analysis compares the work specified in the seed with the trajectory and traces omitted work to initial materials, object relationships, or tool behavior.
For example, if a task asks the agent to implement a core function that the environment already provides, the remaining work may reduce to changing a single constant.
Reconstruction restores the need to implement and debug the function while retaining the public specifications needed to do so.
The review distinguishes efficient solutions that perform the required work from work already completed during construction, targeting revisions at the latter.
Using the seed $z=z_{b,i}$, trajectory, and diagnosis, the Construction Agent revises the task:
\begin{equation}
  T_{r+1} = \operatorname{Refine}(T_r;z,\tau_r,d_r).
  \label{eq:short-solution-redesign}
\end{equation}
$\operatorname{Refine}$ preserves the goal while revising the initial state, object relationships, implementation requirements, or tool feedback and updating the instruction, reference solution, and verifier accordingly.
The revised task returns to independent execution to check whether the restored requirements enter the actual solving process, closing the task-level challenge--reconstruction loop.

To inform subsequent planning, tasks that pass review and runtime checks form $\mathcal A_b$, while research, execution, review, and reconstruction records provide new feedback $\Delta\mathcal F_b$:
\begin{equation}
  \mathcal D_{b+1} = \mathcal D_b \cup \mathcal A_b,
  \qquad
  \mathcal F_{b+1} = \mathcal F_b \cup \Delta\mathcal F_b.
  \label{eq:batch-archive-update}
\end{equation}
The Planner uses the updated collection and records in the next batch to identify repeated work and decide which behavioral patterns to target next.
Gaps found during execution thus inform both the current task's reconstruction and future research briefs, closing the batch-level outer loop.

\subsection{Progressive Horizon Expansion}
\label{sec:ascent}
A solvable task is not the endpoint of our environment construction, but a starting point for exploring the professional problem more deeply.
We extend accepted tasks with new requirements under the same goal, linking the added work to earlier judgments, actions, and results so that agentic behavioral patterns can continue across stages.

\noindent\textbf{Progressive Content Expansion.}
The Evolution Agent examines an accepted parent task $T^{(g)}$ for valuable extensions under the same goal and forms a proposal $\delta^{(g)}$.
It then checks whether the parent's solution can directly satisfy the new requirements, identifying the further analysis, implementation, or verification needed.
We consider four content-expansion directions, including \textbf{broader roles}, using existing artifacts or states to support related responsibilities, and \textbf{deeper mechanisms}, making previously simplified relationships, dependencies, or uncertainties part of the work.
\textbf{Richer evidence} adds records, observations, and professional materials that inform judgments and cross-source analysis, while \textbf{method upgrades} require substantial extensions or redesign of the original approach.

\noindent\textbf{Implementing Interaction Dependencies.}
The Evolution Agent selects from three interaction designs to connect the added requirements to evidence, artifacts, and states encountered or produced during the task, so that later work depends on earlier progress.
To support \textbf{exploration and judgment}, it makes later decisions depend on evidence gathered through public queries, probes, and observations.
In \textbf{action and feedback}, observable tool or system effects link earlier operations to subsequent diagnosis, correction, and verification.
With \textbf{instance variation and adaptation}, objects or states vary across runs under a fixed public specification, requiring the agent to inspect the current instance, adapt its actions, and verify the results.
These designs are selected to serve the extension, not to impose a fixed sequence, and are implemented with the content proposal as a new candidate task:
\begin{equation}
  \widetilde T^{(g+1)}
  = \operatorname{Expand}\!\left(T^{(g)};\delta^{(g)}\right).
  \label{eq:content-interaction-evolution}
\end{equation}
$\operatorname{Expand}$ produces a separate candidate $\widetilde T^{(g+1)}$ whose public specification combines the original and added requirements.
The environment, reference solution, and verifier are updated accordingly, and new scoring conditions continue to follow the public completion contract in \eqref{eq:public-contract}.

An accepted candidate becomes $T^{(g+1)}$, the parent task for the next expansion.

\begin{figure}[t]
    \centering
    \includegraphics[width=\linewidth]{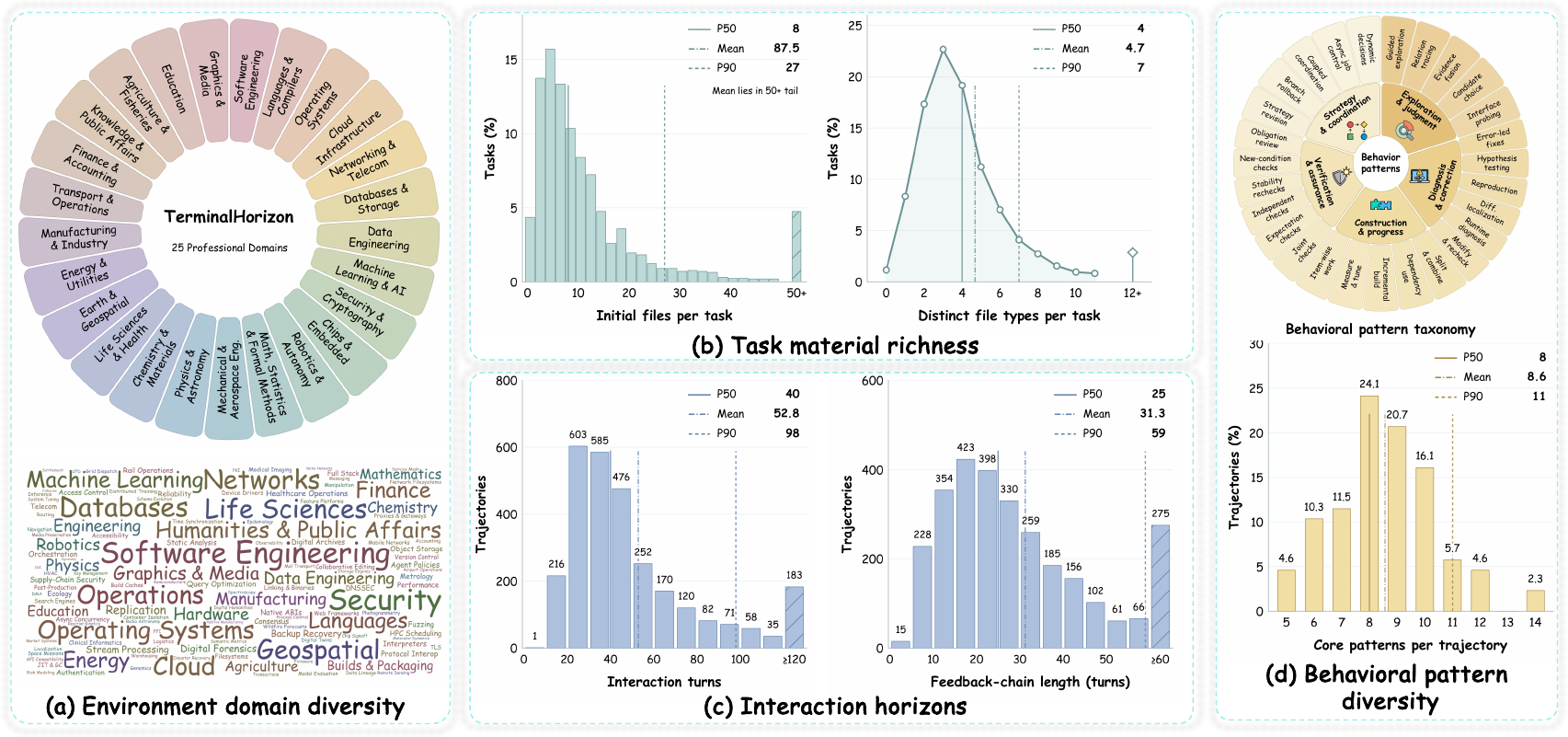}
    \vspace{-5ex}
    \caption{
        Comprehensive analysis of task environments and trajectories in TerminalHorizon.
    }
    \label{fig:data}
    \vspace{-4ex}
\end{figure}

\section{Analysis of Synthesized Data}
\label{Synthesized Data}
The following analysis illustrates the training experience provided by TerminalHorizon: our model can learn not only how to execute terminal commands but, more importantly, how to organize sustained work toward a goal, observe feedback, revise strategies, and deliver final results.

\noindent\textbf{Trajectory Collection.}
We use TerminalHorizon to automatically generate 1,535 task environment packages and collect two trajectories per task using DeepSeek-V4-Pro-0813 as the teacher model and Terminus2 as the execution harness.
We then apply the three filtering criteria described in Appendix~\ref{app:trajectory_filtering}, retaining 2,852 trajectories to form TerminalHorizon-3K.
We conduct rigorous benchmark decontamination before training (Appendix~\ref{app:benchmark_decontamination}).

\noindent\textbf{Broad coverage of real-world domains.}
As shown in Figure~\hyperref[fig:data]{\ref*{fig:data}a}, TerminalHorizon-3K brings real-world work in areas such as chips, physics, and astronomy into a unified terminal interaction format, covering 25 professional domains.
This diversity is enabled by the Planner Agent's scheduling and coordination throughout the automated construction process. See Appendix~\ref{app:professional-domain-coverage} for details.

\begin{wrapfigure}[11]{r}{0.37\textwidth}
  \vspace{-3ex}
  \centering
  \includegraphics[width=\linewidth]{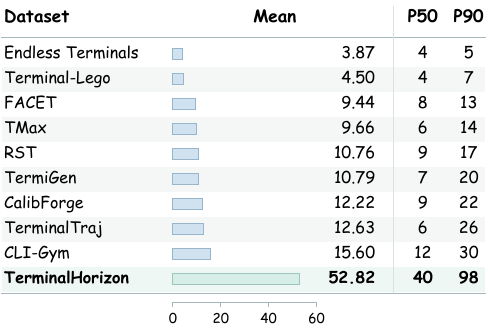}
  \setlength{\abovecaptionskip}{4pt}
  \setlength{\belowcaptionskip}{0pt}
  \vspace{-4ex}
  \caption{Interaction turns per trajectory: mean, P50, and P90.}
  \vspace{-3ex}
  \label{fig:interaction-turns-comparison}
\end{wrapfigure}

\noindent\textbf{Sustained interaction.}
For a fair comparison, we randomly sample 300 tasks from each of several open-source terminal-task synthesis methods and run experiments under the same teacher protocol and settings.
Our trajectories have a mean interaction length of 52.8 turns and a 90th percentile of 98 turns, substantially exceeding those of other methods (Figure~\ref{fig:interaction-turns-comparison}).
We also measure the longest continuous action chain linked by terminal feedback to examine how long feedback remains part of the interaction.
Figure~\hyperref[fig:data]{\ref*{fig:data}c} shows a median chain length of 25 turns.
These results show that longer execution includes sustained sequences of actions and environment feedback, demonstrating how to connect observations with subsequent actions across multiple turns.

\noindent\textbf{Agentic behavioral patterns beyond individual commands.}
\label{agentic patterns}
Beyond interaction length, we examine how these actions organize problem solving.
As shown in Figure~\hyperref[fig:data]{\ref*{fig:data}d}, we group 27 behavioral patterns into five categories: information exploration and judgment, diagnosis and error correction, construction and work progress, verification and assurance, and strategy and execution coordination.
These patterns describe how agents organize multi-step actions into a problem-solving process.
For example, subproblem decomposition and integration breaks a task into subproblems and combines their results to achieve a shared goal.
Detailed pattern definitions are provided in Appendix~\ref{app:agentic-behavioral-patterns}.
Our trajectories contain an average of 8.6 core patterns, providing diverse behavioral experience spanning exploration, construction, error correction, and verification.


\begin{table}[t]
  \centering
  \setlength{\belowcaptionskip}{6pt}
  \caption{Main results on Terminal-Bench 1.0 and 2.0. \textbf{Bold} marks the best score in each benchmark. - denotes unavailable scores, and $\dagger$ denotes results from official model cards.}
  \label{tab:main-results}
  \begingroup
    \fontsize{8.5}{11}\selectfont
    \setlength{\tabcolsep}{2pt}
    \renewcommand{\arraystretch}{1.10}
    \setlength{\heavyrulewidth}{0.8pt}
    \setlength{\lightrulewidth}{0.35pt}
    \arrayrulecolor{black!65}

    \begin{tabularx}{\linewidth}{
      >{\raggedright\arraybackslash}X
      >{\raggedright\arraybackslash}p{76pt}
      >{\centering\arraybackslash}p{25pt}
      >{\centering\arraybackslash}p{57pt}
      >{\centering\arraybackslash}p{28.5pt}
      >{\centering\arraybackslash}p{28.5pt}}
      \toprule
      \textbf{Model}
        & \textbf{Base Model}
        & \textbf{Traj.}
        & \textbf{Scaffold}
        & \textbf{TB 1.0}
        & \textbf{TB 2.0} \\
      \midrule

      \rowcolor{nblue!6}
      \multicolumn{6}{l}{
        \textcolor{nblue!65!black}{
          \bfseries\itshape Terminal task synthesis methods
        }
      } \\

      \rowcolor{white}
      \rowcolor{black!2.5}
      \mbox{CLI-Universe-8B~\citep{DBLP:journals/corr/abs-2606-22883}}
        & Qwen3-8B & 6k & Terminus2 & 19.1 & 10.9 \\

      \mbox{TerminalTraj-14B~\citep{wu2026large}}
        & Qwen2.5-Coder-14B & 50.7k & Terminus2 & 28.9 & 19.1 \\

      \rowcolor{black!2.5}
      \mbox{TermiGen-32B~\citep{DBLP:journals/corr/abs-2602-07274}}
        & Qwen2.5-Coder-32B & 3.3k & BashAgent & -- & 19.3 \\

      \mbox{LiberCoder-32B~\citep{DBLP:journals/corr/abs-2602-10999}}
        & Qwen3-32B & 48.3k & OpenHands & 38.9 & 19.5 \\

      \rowcolor{black!2.5}
      \mbox{SkillSynth-14B~\citep{DBLP:journals/corr/abs-2604-25727}}
        & Qwen3-14B & 10.7k & Terminus2 & 22.9 & 19.9 \\

      \mbox{Nemotron-Terminal-14B~\citep{DBLP:journals/corr/abs-2602-21193}}
        & Qwen3-14B & 490.5k & Terminus2 & -- & 20.2 \\

      \rowcolor{black!2.5}
      \mbox{TerminalTraj-32B~\citep{wu2026large}}
        & Qwen2.5-Coder-32B & 50.7k & Terminus2 & 35.3 & 22.0 \\

      \mbox{CLI-Universe-14B~\citep{DBLP:journals/corr/abs-2606-22883}}
        & Qwen3-14B & 6k & Terminus2 & 27.5 & 23.0 \\

      \rowcolor{black!2.5}
      \mbox{Terminal-Lego-32B~\citep{DBLP:journals/corr/abs-2606-03461}}
        & Qwen3-32B & 15.4k & Terminus2 & -- & 24.3 \\

      \mbox{OpenThinker-Agent-32B~\citep{DBLP:journals/corr/abs-2606-24855}}
        & Qwen3-32B & 100k & Terminus2 & -- & 26.2 \\

      \rowcolor{black!2.5}
      \mbox{Nemotron-Terminal-32B~\citep{DBLP:journals/corr/abs-2602-21193}}
        & Qwen3-32B & 490.5k & Terminus2 & -- & 27.4 \\

      \mbox{SkillSynth-32B~\citep{DBLP:journals/corr/abs-2604-25727}}
        & Qwen3-32B & 10.7k & Terminus2 & 33.8 & 29.6 \\

      \rowcolor{black!2.5}
      \mbox{LiberCoder-235B~\citep{DBLP:journals/corr/abs-2602-10999}}
        & Qwen3-235B-A22B & 48.3k & OpenHands & 46.1 & 31.0 \\

      \mbox{Terminal-World-32B~\citep{DBLP:journals/corr/abs-2605-20876}}
        & Qwen3-32B & 5.7k & Terminus2 & -- & 31.5 \\

      \rowcolor{black!2.5}
      \mbox{Meta-Task-32B~\citep{DBLP:journals/corr/abs-2607-27929}}
        & Qwen3-32B & 3.2k & Terminus2 & -- & 31.8 \\

      \mbox{CLI-Universe-32B~\citep{DBLP:journals/corr/abs-2606-22883}}
        & Qwen3-32B & 6k & Terminus2 & 38.8 & 33.4 \\

      \rowcolor{black!2.5}
      \mbox{TMax-27B~\citep{DBLP:journals/corr/abs-2606-23321}}
        & Qwen3.6-27B & 16.5k & Vanillux2Agent & -- & 42.7 \\

      \mbox{CalibForge-35B-A3B~\citep{DBLP:journals/corr/abs-2608-06352}}
        & Qwen3.5-35B-A3B & 5.4k & CalibForge-Eval & -- & 47.6 \\

      \rowcolor{black!2.5}
      \mbox{RST-27B~\citep{RST}}
        & Qwen3.5-27B & 327.2k & Terminus2 & -- & 49.4 \\

      \mbox{Terminal-Universe-27B~\citep{wu2026terminaluniverseturningagenttrajectories}}
        & Qwen3.5-27B & 32.0k & Terminus2 & -- & 52.8 \\

      \midrule
      \rowcolor{green!5}
      \multicolumn{6}{l}{
        \textcolor{darker}{\bfseries\itshape Ours}
      } \\

      \rowcolor{white}
      \mbox{Qwen3.5-27B {\footnotesize\textcolor{black!60}{(base)}}}
        & -- & -- & Terminus2 & 46.3 & 41.6$^{\dagger}$ \\

      \rowcolor{green!8}
      \mbox{\textcolor{darker}{\textbf{TerminalHorizon-27B}}}
        & Qwen3.5-27B & 3k & Terminus2
        & \textbf{55.0} & \textbf{55.8} \\

      \bottomrule
    \end{tabularx}
    \vspace{-5ex}
  \endgroup
\end{table}

\section{Experiments}
\label{sec:exp}

\subsection{Experimental Setup}
\noindent\textbf{Evaluation Benchmarks.}
We evaluate the model on 12 benchmarks spanning terminal task solving, real-world productivity, and general reasoning.
\textbf{(1) Coding:} Terminal-Bench 1.0, 2.0, and 2.1~\citep{Terminal-Bench} assess terminal task-solving performance.
\textbf{(2) Agentic:} AutomationBench~\citep{shepard2026automationbench}, the Airline and Retail domains of $\tau^2$-bench~\citep{tau-bench}, and the English Cross-domain main set of VitaBench~\citep{VitaBench} test whether experience learned from long-horizon trajectories transfers to real-world productivity tasks.
\textbf{(3) General reasoning:} MATH-500~\citep{math500}, GSM8K~\citep{gsm8k}, AIME 2024~\citep{maa2024aime}, AIME 2025~\citep{maa2025aime}, and KOR-Bench (Cipher)~\citep{KOR-Bench} measure how long-horizon interaction training affects general reasoning ability.
Detailed evaluation settings are provided in Appendix~\ref{app:evaluation-protocols}.

\noindent\textbf{Implementation Details.}
We apply SFT to Qwen3.5-27B on TerminalHorizon-3K to obtain TerminalHorizon-27B.
The maximum sequence length is \(262{,}144\) tokens.
We use fused AdamW with a learning rate of \(1\times10^{-5}\), weight decay of \(0.1\), and \((\beta_1,\beta_2)=(0.9,0.95)\).
The learning rate follows a cosine schedule after a \(5\%\) warmup, and the gradient clipping threshold is \(1.0\).
Training runs for \(3\) epochs in BF16 with DeepSpeed ZeRO-3.
Unless otherwise noted, we use \(144\) H20-141GB GPUs across \(18\) nodes.
The full training configuration is provided in Appendix~\ref{app:training-details}.

\subsection{Main Results}
Tables~\ref{tab:main-results} and~\ref{tab:extended-results} show that TerminalHorizon-27B outperforms its Qwen3.5-27B base model on all 12 benchmarks across three categories.
We discuss these gains from three perspectives.

\noindent\textbf{Stronger performance with fewer data.}
TerminalHorizon-27B achieves 55.0\%, 55.8\%, and 60.67\% on Terminal-Bench 1.0, 2.0, and 2.1, respectively, improving over the base model by 8.7, 14.2, and 13.85\%.
As shown in Table~\ref{tab:main-results}, it also outperforms Terminal-Universe-27B by 3.0\% using the same base model and harness but fewer than one-tenth as many training trajectories.
This result demonstrates the data efficiency of long-horizon experience and highlights the high information density of our trajectories, which are constructed around sustained problem solving.


\begin{wraptable}[21]{r}{0.51\textwidth}
\centering
\begin{threeparttable} 

\setlength{\belowcaptionskip}{2pt}
\caption{Performance across coding, agentic, and general reasoning benchmarks.}
\label{tab:extended-results}
\begingroup
\fontsize{8}{10}\selectfont
\setlength{\tabcolsep}{2pt}
\renewcommand{\arraystretch}{1.10}
\setlength{\heavyrulewidth}{0.8pt}
\setlength{\lightrulewidth}{0.35pt}
\arrayrulecolor{black!65}

\begin{tabular}{
  @{}l@{\hspace{4pt}}c
  >{\columncolor{green!5}[\tabcolsep][0pt]}c@{}}
  \toprule
  \textbf{Benchmark}
    & \begin{tabular}[c]{@{}c@{}}
        \textbf{Qwen3.5-}\\[-1pt]
        \textbf{27B}
      \end{tabular}
    & \begin{tabular}[c]{@{}c@{}}
        \textcolor{darker}{\textbf{TerminalHorizon-}}\\[-1pt]
        \textcolor{darker}{\textbf{27B}}
      \end{tabular} \\
  \midrule

  \rowcolor{nblue!6}[0pt][0pt]
  \multicolumn{3}{@{}l@{}}{
    \textcolor{nblue!65!black}{\bfseries\itshape Coding}
  } \\
  Terminal-Bench 2.1 & 46.82
    & \textbf{60.67}\,{\fontsize{7.5}{9}\selectfont
      \textcolor{darker}{(+13.85)}} \\
  \midrule

  \rowcolor{nblue!6}[0pt][0pt]
  \multicolumn{3}{@{}l@{}}{
    \textcolor{nblue!65!black}{\bfseries\itshape Agentic}
  } \\
  AutomationBench (Pass Rate) & 5.5
    & \textbf{17.2}\,{\fontsize{7.5}{9}\selectfont
      \textcolor{darker}{(+11.7)}} \\
  AutomationBench (Partial) & 26.4
    & \textbf{49.7}\,{\fontsize{7.5}{9}\selectfont
      \textcolor{darker}{(+23.3)}} \\
  $\tau^2$-bench Airline & 79.0
    & \textbf{84.5}\,{\fontsize{7.5}{9}\selectfont
      \textcolor{darker}{(+5.5)}} \\
  $\tau^2$-bench Retail & 84.9
    & \textbf{87.5}\,{\fontsize{7.5}{9}\selectfont
      \textcolor{darker}{(+2.6)}} \\
  VitaBench Cross-domain & 20.5
    & \textbf{31.8}\,{\fontsize{7.5}{9}\selectfont
      \textcolor{darker}{(+11.3)}} \\
  \midrule

  \rowcolor{nblue!6}[0pt][0pt]
  \multicolumn{3}{@{}l@{}}{
    \textcolor{nblue!65!black}{\bfseries\itshape General reasoning}
  } \\
  MATH-500 & 96.9
    & \textbf{98.9}\,{\fontsize{7.5}{9}\selectfont
      \textcolor{darker}{(+2.0)}} \\
  GSM8K & 96.3
    & \textbf{97.3}\,{\fontsize{7.5}{9}\selectfont
      \textcolor{darker}{(+1.0)}} \\
  AIME 2024 & 93.3
    & \textbf{96.7}\,{\fontsize{7.5}{9}\selectfont
      \textcolor{darker}{(+3.4)}} \\
  AIME 2025 & 86.7
    & \textbf{93.3}\,{\fontsize{7.5}{9}\selectfont
      \textcolor{darker}{(+6.6)}} \\
  KOR-Bench (Cipher) & 90.1
    & \textbf{91.2}\,{\fontsize{7.5}{9}\selectfont
      \textcolor{darker}{(+1.1)}} \\
  \bottomrule
\end{tabular}

\arrayrulecolor{black}
\endgroup
\end{threeparttable} 

\end{wraptable}

\noindent\textbf{Transfer to real-world productivity tasks.}
Training on our high-quality long-horizon trajectories also improves performance on real-world enterprise automation, customer service, and service tasks spanning multiple scenarios, with gains across all four agentic benchmarks. 
On AutomationBench, Pass Rate and Partial scores improve by 11.7\% and 23.3\%, respectively.
VitaBench improves by 11.3\%, while $\tau^2$-bench Airline and Retail improve by 5.5\% and 2.6\%, respectively.
These results show that our training benefits extend beyond terminal operations, enabling the model to reliably handle complex environmental feedback throughout real-world long-horizon tasks and apply this capability consistently across tasks.

\noindent\textbf{General reasoning is preserved alongside agentic gains.}
Performance on the five general reasoning benchmarks is preserved and shows further gains.
MATH-500 reaches 98.9\%, while AIME 2024 and AIME 2025 reach 96.7\% and 93.3\%, improving over the base model by 3.4\% and 6.6\%, respectively.
GSM8K and KOR-Bench also improve by 1.0\% and 1.1\%.
Thus, long-horizon training improves agentic task performance while preserving and strengthening general reasoning.

\subsection{Analysis and Ablation Studies}


\suppressfloats[t]
\begin{figure}[t]
  \centering
  \setlength{\abovecaptionskip}{0pt}
  \setlength{\belowcaptionskip}{0pt}
  \begin{minipage}[t]{0.49\linewidth}
    \vspace{0pt}
    \parbox[b][110pt][b]{\linewidth}{\includegraphics[width=\linewidth]{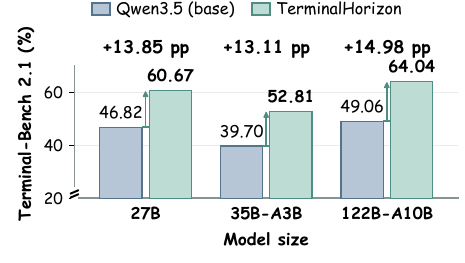}}
    \caption{TB~2.1 performance across model sizes.}
    \label{fig:model-size-ablation}
  \end{minipage}\hfill
  \begin{minipage}[t]{0.48\linewidth}
    \vspace{0pt}
    \parbox[b][110pt][b]{\linewidth}{\includegraphics[width=\linewidth]{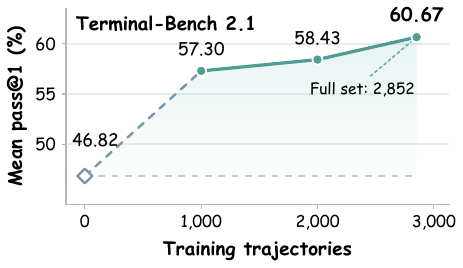}}
    \caption{Training-data scaling on TB~2.1.}
    \label{fig:training-data-scaling}
  \end{minipage}
  \vspace{-2ex}
\end{figure}
\leavevmode

\begin{wraptable}[10]{r}{0.39\textwidth}
  \vspace{-3ex}
  \raggedleft
\begingroup
  \makeatletter\def\@captype{table}\makeatother
  \setlength{\abovecaptionskip}{0pt}
  \setlength{\belowcaptionskip}{3pt}
  \begin{threeparttable}
    \caption{Interaction statistics and task success with two teacher models.}
    \label{tab:teacher-comparison}
    \fontsize{8.5}{10.5}\selectfont
    \setlength{\tabcolsep}{2pt}
    \renewcommand{\arraystretch}{1.05}
    \setlength{\heavyrulewidth}{0.7pt}
    \setlength{\lightrulewidth}{0.35pt}
    \arrayrulecolor{black!60}
    \begin{tabular}{@{}lcc@{}}
      \toprule
      \rowcolor{nblue!6}
      \textbf{Metric}
        & \textbf{GLM-5.2}
        & \begin{tabular}[c]{@{}c@{}}
            \textcolor{darker}{\textbf{DeepSeek-}}\\[-1pt]
            \textcolor{darker}{\textbf{V4-Pro}}
          \end{tabular} \\
      \midrule
      Mean turns & 46.1 & 53.5 \\
      Mean feedback chain & 19.2 & 30.3 \\
      Exact repeats (\%) & 15.8 & 13.2 \\
      \midrule
      \rowcolor{darker!6}
      \textbf{Pass rate (\%)} & \textbf{44.0} & \textbf{50.0} \\
      \bottomrule
    \end{tabular}
    \arrayrulecolor{black}
  \end{threeparttable}
\endgroup
  \vspace{-3ex}
\end{wraptable}

\noindent\textbf{Effect of Teacher Model.}
We randomly sample 100 of the 1,535 tasks to compare interaction characteristics between two teacher models, GLM-5.2 and DeepSeek-V4-Pro-0813.
As shown in Table~\ref{tab:teacher-comparison}, both teachers generate long trajectories on average, indicating that our tasks can provide long-horizon problem-solving experience across teachers.
DeepSeek-V4-Pro produces longer feedback chains on average and has a lower rate of exact action repetition.
These more sustained interactions are also associated with a higher task completion rate, supporting its selection as our teacher for collecting long-horizon trajectories.

\noindent\textbf{Performance across Model Sizes.}
To examine whether the training benefits of TerminalHorizon extend across model sizes, we also evaluate its effectiveness on Qwen3.5-35B-A3B and Qwen3.5-122B-A10B.
As shown in Figure~\ref{fig:model-size-ablation}, all three configurations achieve double-digit percentage-point gains over their base models, improving by 13.85\%, 13.11\%, and 14.98\%.
These results show that the agentic behavioral patterns captured in our long-horizon trajectories support improvements across models, so their training value is not limited to a single base model.

\begin{samepage}
\noindent\textbf{Training Data Scaling.}
As shown in Figure~\ref{fig:training-data-scaling}, scaling from 1000 to approximately 3k training trajectories raises performance on Terminal-Bench 2.1 to 60.67\%.
This demonstrates that data generated by TerminalHorizon can effectively scale agent capabilities.
We also observe no sign of performance saturation, suggesting that further expanding high-quality long-horizon training trajectories may yield additional gains.
\par
\end{samepage}

\subsection{Long-Horizon Task Applications}

\begin{figure}[t]
    \centering
    \includegraphics[width=\linewidth]{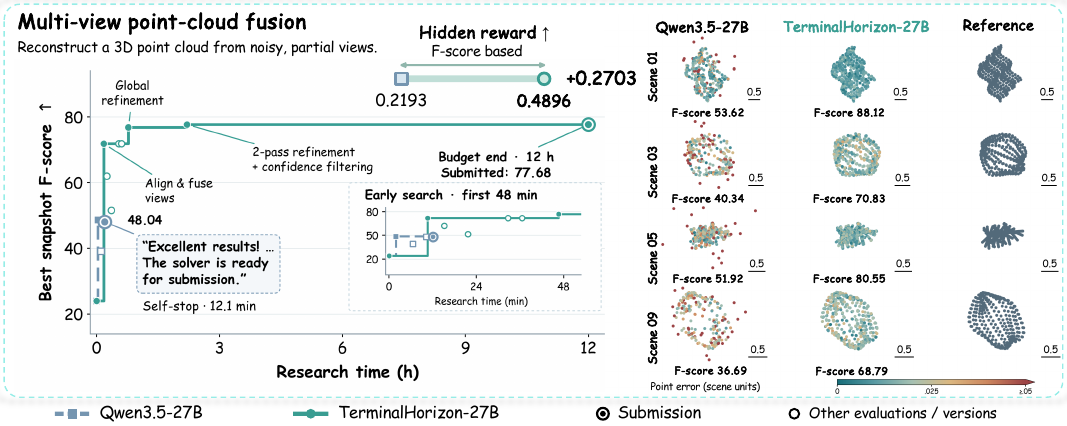}
    \vspace{-4ex}
\caption{Long-horizon optimization on RSI-Exam.
More cases in Appendix~\ref{app:additional-long-horizon-applications}.}
    \label{fig:long-horizon-applications}
    \vspace{-4ex}
\end{figure}

Autonomous long-horizon tasks require agents to continue seeking improvements after finding an initial solution, revise their methods based on feedback, and carry effective progress through to final delivery.
To evaluate this system-level capability, we compare TerminalHorizon-27B with Qwen3.5-27B on several RSI-Exam cases under the same 12-hour autonomous research budget.
Taking multi-view point cloud fusion as an example, Figure~\ref{fig:long-horizon-applications} shows both models' optimization processes and final results.
The base model obtains a high score on a self-designed proxy metric and voluntarily stops after just 12.1 minutes, judging its results to be excellent and ready for submission.
However, agreement with a single reference view does not ensure reconstruction of the complete scene: a method can perform well on this proxy while missing surfaces visible only in other views.

In contrast, our model continues experimenting beyond the initial solution and revisits both its fusion method and the measurements used to assess progress.
It inspects the contributions of different views and introduces a leave-one-view-out diagnostic, which exposes insufficient coverage and guides subsequent revisions.
It then improves cross-view alignment, performs global refinement, and uses confidence-based point selection to improve the reconstruction.
These changes reflect substantive revisions informed by experimental evidence, rather than simply repeating the same operations for longer.
Our model retains the resulting improvements in its final submission, whose hidden evaluation score exceeds the base model's by 27.03\%.
This comparison illustrates the practical value of experience with agentic behavioral patterns: connecting exploration, diagnosis, method revision, and verification enables agents to keep seeking improvements and turn verified progress into final deliverables.
Additional cases are provided in Appendix~\ref{app:additional-long-horizon-applications}.
\section{Related Work}
\label{sec:related}

\noindent\textbf{Long-Horizon Agents.}
Long-horizon agents must retain information, track progress, and coordinate actions across stages~\citep{dong2026longhorizon,SaaSBench,Last-Exam}.
Harnesses support execution through context, memory, and state management, helping maintain continuity during extended interactions~\citep{AgentFold,ma2026longhorizon,ReAcTree}.
We complement this support by studying how training develops sustained problem-solving ability, particularly the agentic behavioral patterns that connect exploration, implementation, and verification around a shared goal.

\noindent\textbf{Terminal Environment Synthesis.}
Environment synthesis scales interaction experience through programmatic construction and evolution~\citep{EnvScaler,Agent-World-Model,Agent-World}.
Terminal-task synthesis uses skill composition, technical evidence, and real-world needs~\citep{wu2026terminaluniverseturningagenttrajectories,DBLP:journals/corr/abs-2605-20876,wu2026large,DBLP:journals/corr/abs-2604-25727,FACET,DBLP:journals/corr/abs-2606-22883,DBLP:journals/corr/abs-2602-21193}, while recursive reference-solution expansion increases complexity~\citep{RST}.
Beyond coverage and complexity, we emphasize these patterns' less-studied cross-stage connections: findings inform decisions, feedback guides strategy revision, and intermediate results support delivery.
\section{Conclusion}
\label{conclusion}

We introduced TerminalHorizon, a fully automated engine for synthesizing long-horizon trajectories that connect agentic behavioral patterns.
Through challenge-guided reconstruction and progressive expansion, it links exploration, strategy revision, and verification around a shared goal.
Training on these trajectories improves performance on all 12 benchmarks, while RSI-Exam cases show sustained experimentation, feedback-driven improvement, and progress retained through final delivery.
These findings highlight the training value of long-horizon behavioral experience in advancing agents from local terminal operations toward sustained autonomous work.

\subsection*{AI use statement}
In this work, we used generative AI for automated task environment construction and solution trajectory generation as part of our research methodology, and for language polishing and the generation of selected illustration assets during manuscript preparation. The procedures and validation mechanisms for task environment construction and trajectory generation are described in the Method section and the appendix. We did not use generative AI to formulate research questions, develop the core research ideas, or design the research methodology.
We manually reviewed all AI-assisted content. For language polishing, we checked that the revised wording preserved the intended meaning. For generated illustration assets, we checked that they accurately conveyed the relevant concepts without introducing misleading information.
We take full responsibility for the final content of this work, including text, claims, and illustration assets produced with the aid of generative AI.

\subsection*{Ethics statement}

This work studies long-horizon task learning for terminal agents through synthetic task environments, generated solution trajectories, and benchmark evaluations, examining how sustained interaction experience affects agent capabilities. Within the scope of this study, we have identified no significant ethical concerns requiring specific disclosure. The procedures for task construction, data filtering, and experimental evaluation are detailed in the main text and the appendix.

\subsection*{Reproducibility statement}

To support reproducibility, we describe the task formulation and two-stage synthesis procedure in Section~\ref{sec:method}, and trajectory collection and experimental settings in Section~\ref{Synthesized Data} and Section~\ref{sec:exp}.
Appendix~\ref{app:training-details} provides the training configuration, while Appendix~\ref{app:evaluation-protocols} specifies the benchmark protocols, including agent frameworks, decoding settings, interaction limits, and scoring procedures.
Trajectory filtering and benchmark decontamination are described in Appendices~\ref{app:trajectory_filtering} and~\ref{app:benchmark_decontamination}, with the corresponding prompt templates in Appendix~\ref{app:prompt_templates}.
For the long-horizon RSI case studies, Appendix~\ref{app:additional-long-horizon-applications} details the research setup, evaluation procedure, and evidence underlying the case analyses.



\bibliography{iclr2027_conference}
\bibliographystyle{iclr2027_conference}

\appendix
\clearpage
\section{Implementation Details}
\label{app:IMPLEMENTATION-details}

\subsection{Training Details}
\label{app:training-details}

We use LLaMA-Factory~\citep{LlamaFactory} to perform full-parameter supervised fine-tuning (SFT) on multi-turn trajectories from TerminalHorizon-3K for Qwen3.5-27B, Qwen3.5-35B-A3B, and Qwen3.5-122B-A10B\footnote{\url{https://huggingface.co/collections/Qwen/qwen35}}.
All three models follow the same training recipe in Table~\ref{tab:training-hyperparameters}, and we report the final checkpoint after three epochs.


\begin{table}[!hbp]
  \centering
  \begin{minipage}{0.78\linewidth}
  \setlength{\belowcaptionskip}{6pt}
  \caption{Shared SFT hyperparameters for all three model sizes.}
  \label{tab:training-hyperparameters}
  \begingroup
  \fontsize{9}{11}\selectfont
  \setlength{\tabcolsep}{6pt}
  \renewcommand{\arraystretch}{1.12}
  \setlength{\heavyrulewidth}{0.7pt}
  \setlength{\lightrulewidth}{0.35pt}
  \arrayrulecolor{black!60}
  \begin{tabularx}{\linewidth}{
    >{\raggedright\arraybackslash}X
    >{\raggedright\arraybackslash}X}
    \toprule
    \rowcolor{nblue!6}
    \textbf{Hyperparameter} & \textbf{Value} \\
    \midrule
    Optimizer & AdamW $(\beta_1{=}0.9,\,\beta_2{=}0.95)$ \\
    \rowcolor{black!2.5}
    Peak learning rate & $1.0\times10^{-5}$ \\
    LR scheduler & Cosine \\
    \rowcolor{black!2.5}
    Warmup ratio & $0.05$ \\
    Weight decay & $0.1$ \\
    \rowcolor{black!2.5}
    Epochs & 3 \\
    Maximum sequence length & $262{,}144$ \\
    \rowcolor{black!2.5}
    Sequence packing & Disabled \\
    Per-device train batch size & 1 \\
    \rowcolor{black!2.5}
    Precision & BF16 \\
    Distributed optimization & DeepSpeed ZeRO-3 \\
    \rowcolor{black!2.5}
    Attention backend & FlashAttention-2 \\
    Random seed & 42 \\
    \bottomrule
  \end{tabularx}
  \endgroup
  \end{minipage}
\end{table}

\subsection{Evaluation Protocols}
\label{app:evaluation-protocols}

\paragraph{General settings.}
All evaluated models are served through vLLM~\citep{vllm} on eight H20 GPUs, with a tensor parallelism degree of 8 and a context window of $262{,}144$ tokens.
For each benchmark, the base and TerminalHorizon models use the same task set, agent framework, decoding settings, and scoring procedure.
Terminal-Bench, AutomationBench, and general reasoning evaluations use a temperature of $1.0$, top-$p=0.95$, top-$k=20$, and a maximum of $65{,}536$ output tokens per response.
Decoding settings for VitaBench and $\tau^2$-bench are specified below.


\paragraph{Terminal-Bench.}
We evaluate on Terminal-Bench 1.0, 2.0, and 2.1~\citep{Terminal-Bench} using Harbor\footnote{\url{https://github.com/harbor-framework/harbor}} and Terminus-2 with a JSON parser.
Terminal-Bench 1.0 contains 80 tasks and uses Daytona sandboxes, each allocated 2 CPUs, 4\,GiB of memory, and 8\,GiB of storage.
Terminal-Bench 2.0 and 2.1 each contain 89 tasks and use E2B sandboxes.
All three versions allow the agent three hours per task, while verifiers retain the time limits specified by each task.
Terminus-2 triggers proactive summarization when the estimated remaining context capacity falls below $131{,}072$ tokens.
We run each task independently three times and report mean pass@1, with up to two retries for infrastructure errors.


\paragraph{AutomationBench.}
We evaluate AutomationBench v1.0.6~\citep{shepard2026automationbench} with 600 rollouts over the full public set, using the official \texttt{api} tool set and a limit of 50 steps per trajectory.
Each model is evaluated in three independent runs, and we report the mean official Partial score (the fraction of assertions passed) and Pass Rate (all assertions passed).
Interrupted samples are retried for up to three rounds; samples that remain incomplete receive a score of zero.


\paragraph{VitaBench Cross-domain.}
We use the main English Cross-domain set of VitaBench 1.0~\citep{VitaBench}, which contains 100 tasks spanning food delivery, in-store services, and travel.
Each task is run four times with a limit of 300 interaction steps, an error limit of 10, and a random seed of 300.
The agent uses a temperature of 0 and a maximum of $65{,}536$ output tokens per response.
The user simulator and trajectory judge both use \texttt{deepseek-v4-pro-0813}, with a temperature of 0 and an output limit of $32{,}768$ tokens.
We use the official rubric-based sliding-window trajectory evaluator and report Avg@4, the mean success rate over all 400 runs.


\paragraph{$\tau^2$-bench.}
We evaluate the Airline (50 tasks) and Retail (114 tasks) domains on the base split of $\tau^2$-bench v1.0.1~\citep{tau-bench}.
We use the official \texttt{llm\_agent} and user simulator, with four independent runs per task, a limit of 200 interaction steps, an error limit of 10, and a random seed of 300.
The agent, user simulator, and natural-language assertion judge all use a temperature of 0; the latter two use \texttt{deepseek-v4-pro-0813}.
We do not specify an additional per-response output limit for this evaluation.
For each domain, we report the mean task success rate across all repeated runs.
Model or API errors are retried up to three times, and these technical retries do not count as additional trials.


\paragraph{General reasoning.}
We evaluate on the 500 problems in MATH-500~\citep{math500}, the 1,319 test problems in GSM8K~\citep{gsm8k}, the 30 problems each in AIME 2024~\citep{maa2024aime} and AIME 2025~\citep{maa2025aime}, and the 250 problems in KOR-Bench Cipher~\citep{KOR-Bench}.
The first four benchmarks use the native evaluators in EvalScope 1.11.1\footnote{\url{https://github.com/modelscope/evalscope}}, while KOR-Bench uses its official zero-shot prompts and answer normalization procedure.
For AIME 2024 and AIME 2025, we use the ModelScope datasets
\texttt{evalscope/aime24}\footnote{\url{https://modelscope.cn/datasets/evalscope/aime24}} and \texttt{evalscope/aime25}\footnote{\url{https://modelscope.cn/datasets/evalscope/aime25}},
respectively, through EvalScope 1.11.1, evaluating all 30
problems in the test split of the default
subset for each year.
We generate one answer per problem and score it according to each benchmark's answer-matching rules, reporting mean accuracy over three independent runs.


\section{Additional Experimental Analysis}
\label{app:additional-experimental-analysis}

\subsection{Performance across Task Categories}
\label{app:task-category-analysis}

\begin{figure}[!ht]
  \centering
  \includegraphics[width=0.59\linewidth]{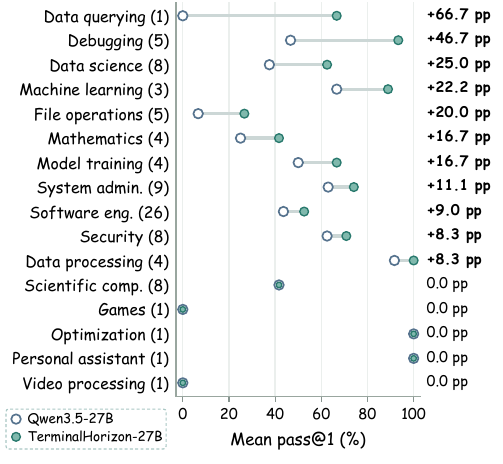}
  \caption{Category-wise mean pass@1 on Terminal-Bench 2.1.}
  \label{fig:tb21-category}
\end{figure}

We compare category-wise mean pass@1 on Terminal-Bench 2.1 to examine how the training benefits are distributed across professional tasks.
As shown in Figure~\ref{fig:tb21-category}, TerminalHorizon-27B matches or outperforms Qwen3.5-27B in all 16 categories, with improvements in 11.
Gains are particularly large in debugging, data science, and machine learning, reaching 46.7, 25.0, and 22.2 percentage points, respectively, and also extend to system administration, security, and data processing.
This distribution shows that the improvements are not concentrated in a single task type but extend across a broad range of professional problem-solving tasks.


\subsection{Data Efficiency}
\label{app:data-efficiency}

\begin{figure}[!ht]
  \centering
  \includegraphics[width=0.95\linewidth]{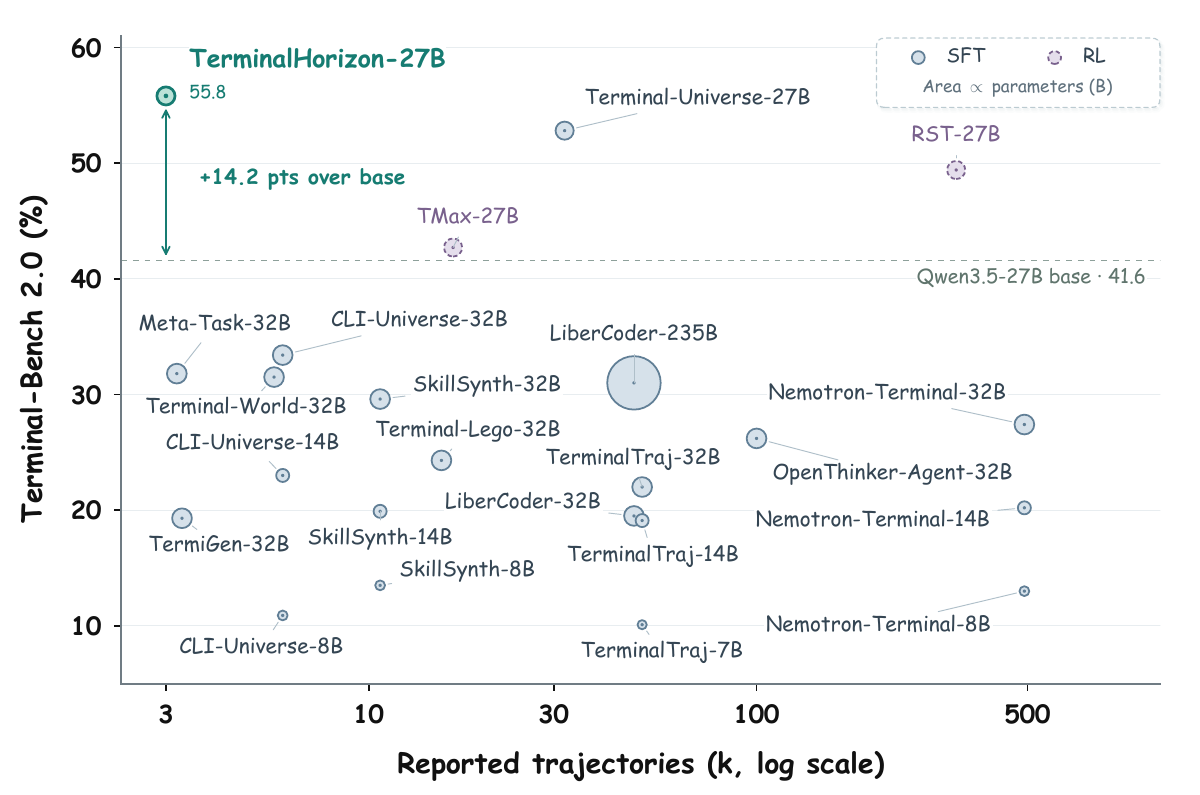}
  \caption{Data efficiency on Terminal-Bench 2.0.
  TerminalHorizon-27B achieves 55.8\% with approximately 3k training trajectories.}
  \label{fig:tb20-efficiency}
\end{figure}

Figure~\ref{fig:tb20-efficiency} compares Terminal-Bench 2.0 performance against the reported number of training trajectories.
With 2,852 trajectories (approximately 3k), TerminalHorizon-27B achieves 55.8\%, the highest score among the models shown.
In particular, it outperforms Terminal-Universe-27B by 3.0 percentage points while using fewer than one-tenth as many trajectories (approximately 3k versus 32k).
Both models use the Qwen3.5-27B base model and Terminus-2 harness, providing a more closely matched comparison within a plot that includes different model sizes and training methods.
These results support the data efficiency of long-horizon experience organized around sustained problem solving, which enables strong terminal-task performance with a limited number of training trajectories.


\section{Additional Long-Horizon Task Applications}
\label{app:additional-long-horizon-applications}
We examine four RSI-Exam\footnote{\url{https://github.com/aiming-lab/RSI-Exam}} case studies to understand how terminal-task training transfers to sustained autonomous research.
The cases span combinatorial optimization, physical inverse problems, game strategy, and geometric reconstruction.
Their shared challenge is not merely to produce an executable solution, but to identify opportunities for further improvement, design informative experiments, and preserve effective changes in the final submission.
These cases complement the aggregate benchmark results by showing how agents organize their work around an ongoing objective.


For each case, we compare Qwen3.5-27B and TerminalHorizon-27B using the same task, research instructions, decoding settings, and evaluation procedure.
Both models operate in an E2B environment with 8\,GiB of memory through a Terminus-2-based research harness, with a research budget of 12 hours.
Agents can inspect the supplied materials, revise their implementations, and run development experiments, but do not receive hidden evaluation results during research.
The task's native verifier scores the final submission rather than a historical checkpoint selected afterward by the evaluator.
Agents may also finish voluntarily before the deadline.


\begin{table}[t]
  \centering
  \begin{minipage}{0.90\linewidth}
  \setlength{\belowcaptionskip}{6pt}
  \caption{RSI-Exam results (12-hour budget; one run per model per case).
  Normalized rewards $\times100$.
  Time excludes setup and final verification.}
  \label{tab:rsi-case-summary}
  \begingroup
  \fontsize{9}{11}\selectfont
  \setlength{\tabcolsep}{4pt}
  \renewcommand{\arraystretch}{1.18}
  \setlength{\heavyrulewidth}{0.7pt}
  \setlength{\lightrulewidth}{0.35pt}
  \arrayrulecolor{black!60}
  \begin{tabularx}{\linewidth}{
    >{\raggedright\arraybackslash}X
    >{\raggedleft\arraybackslash}p{34pt}
    >{\columncolor{darker!6}\raggedleft\arraybackslash}p{34pt}
    >{\raggedleft\arraybackslash}p{34pt}
    >{\raggedleft\arraybackslash}p{34pt}
    >{\raggedleft\arraybackslash}p{34pt}}
    \toprule
    \rowcolor{nblue!6}
    & \multicolumn{3}{c}{\textbf{Hidden set score} $\uparrow$}
    & \multicolumn{2}{c}{\textbf{Research time (h)}} \\
    \rowcolor{nblue!6}
    \multirow{-2}{*}{\textbf{Case}}
    & \textbf{Base} & \textbf{Ours} & \textbf{Gain}
    & \textbf{Base} & \textbf{Ours} \\
    \midrule
    Robot routing and macros & 14.63 & 39.72 & +25.09 & 11.42 & 12.00 \\
    Tidal friction inversion & 27.41 & 48.15 & +20.74 & 1.12 & 12.00 \\
    2048 strategy & 37.54 & 45.25 & +7.71 & 1.28 & 12.00 \\
    Multi-view point-cloud fusion & 21.93 & 48.96 & +27.03 & 0.20 & 12.00 \\
    \bottomrule
  \end{tabularx}
  \endgroup
  \end{minipage}
\end{table}

Table~\ref{tab:rsi-case-summary} summarizes the results for all cases.
Each result corresponds to one research run and is based on the final submission's performance on the hidden test set.
We convert each task's native normalized reward to a 0--100 scale, representing an optimization score rather than a task success rate.
We use the original action--observation traces, experiment records, saved implementations, and final evaluation outputs to analyze how models sustain research and preserve effective improvements.


\subsection{Warehouse Robot Routing and Macro Optimization}
\label{app:rsi-robot}

\begin{figure}[!ht]
  \centering
  \includegraphics[width=\linewidth]{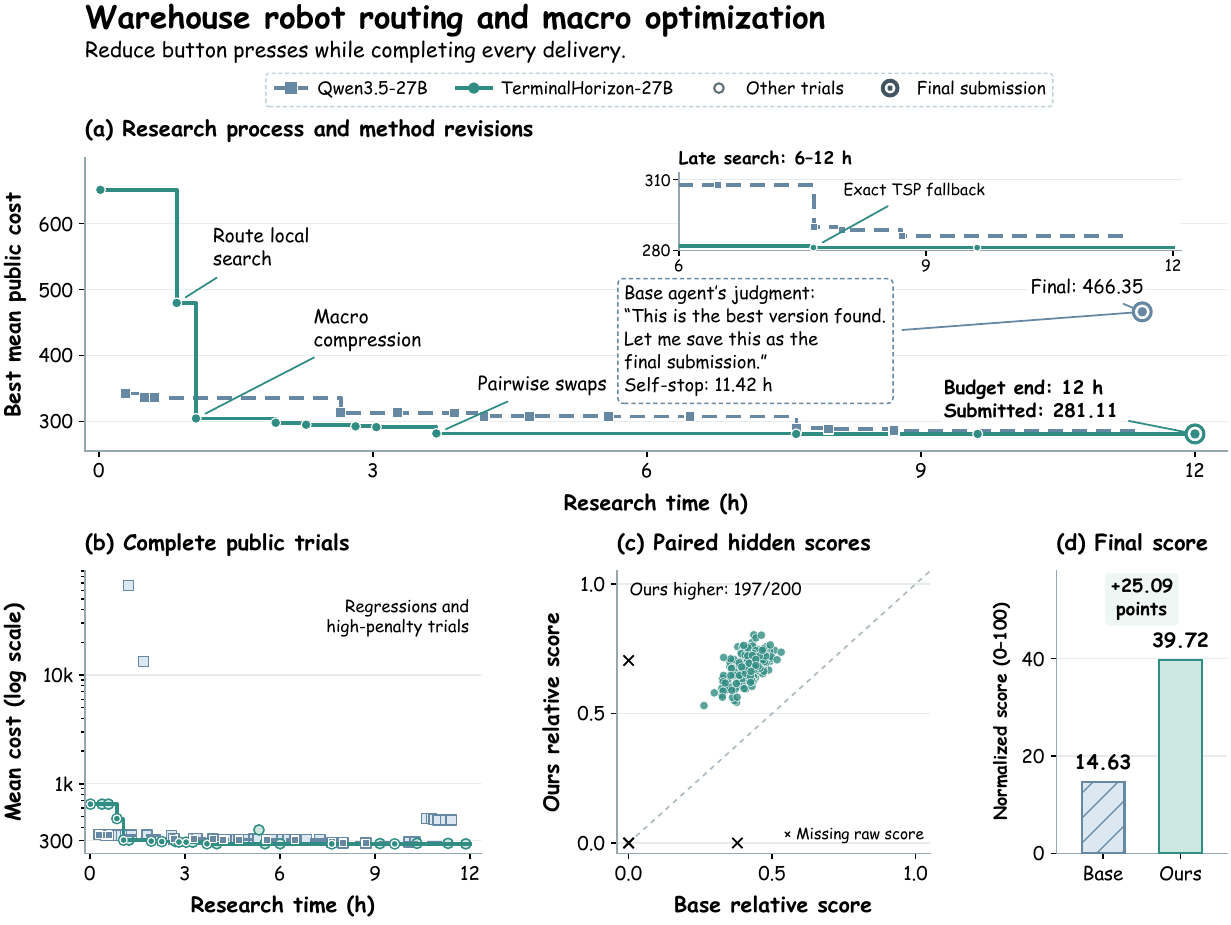}
\caption{Robot routing and macro optimization.
(a) Best mean button cost on 100 public instances. Rings mark final submissions.
(b) All complete public trials, including regressions.
(c) Paired relative scores (reference length / solver cost) on all 200 hidden
instances. Crosses mark missing raw scores, mapped to zero by the grader.
(d) Final normalized scores.}
  \label{fig:app-rsi-robot}
\end{figure}

\paragraph{Task and source of difficulty.}
A robot must collect colored balls in a warehouse with obstacles and deliver each ball to its matching basket.
The controller supports movement, turning, swapping, and the recording and replay of macros.
The goal is to complete all deliveries with as few button presses as possible, subject to a limit on the number of primitive operations after macro expansion.
This couples route planning with program compression: a shorter geometric route need not yield a shorter control program, and macros must remain valid as the robot's position, orientation, and carried object change.
The public evaluator provides 100 instances, while the final verifier uses 200 unseen instances and penalizes incomplete or invalid solutions.
Figure~\ref{fig:app-rsi-robot} connects method revisions with public-set progress and final hidden-set results.


\paragraph{From local improvements to a joint objective.}

TerminalHorizon-27B first builds an orientation-aware route planner, then adds single-macro compression and sequential macro replacement.
Later experiments expand the route search with alternative initializations, route-segment reversals, and pairwise swaps.
On the full public test set, mean cost falls from 651.42 in the initial measured version to 281.92 after pairwise swaps are introduced.
The agent does not accept a candidate merely because it is more complex: it reverts a C++ local-search variant that takes longer and yields a slightly worse cost of 281.96.
In contrast, it retains an exact dynamic-programming route optimizer for smaller instances as an additional candidate.
The final method selects routes by their compressed button count, aligning search with the actual objective rather than uncompressed path length.
The submitted implementation achieves a public mean cost of 281.11.


\paragraph{Preserving progress through final delivery.}

The base model also conducts substantial search, reaching a best recorded cost of 286.05 on the full public set.
However, later revisions remove macro compression, and its actual submission has a cost of 466.35.
It voluntarily stops after 11.42 hours, whereas TerminalHorizon-27B retains its stronger compressed solution through the research deadline.
Their final hidden-set normalized scores are 14.63 and 39.72, respectively.
Both agents discover competitive intermediate solutions, but only TerminalHorizon-27B preserves that progress in the evaluated implementation.
Maintaining a reliable best-so-far solution and checking the final submission are part of long-horizon problem solving, not administrative steps after the substantive work is complete.


\subsection{Tidal Seabed Friction Field Inversion}
\label{app:rsi-tidal}

\begin{figure}[t]
  \centering
  \includegraphics[width=\linewidth]{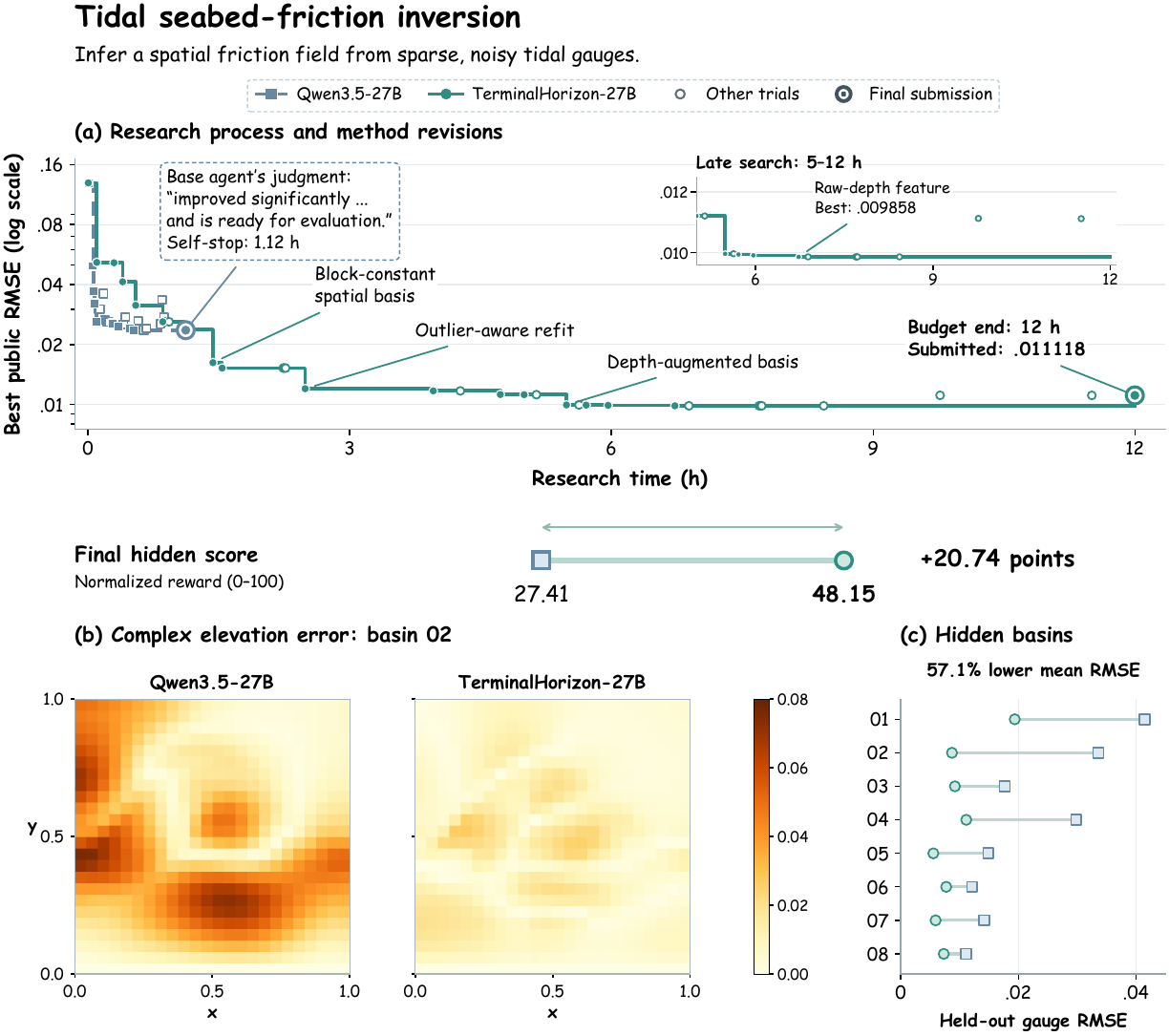}
\caption{Tidal friction inversion.
(a) Best mean public RMSE (six basins).
(b) Unsmoothed $24\times24$ complex-elevation errors with a shared scale.
Basin 02 is selected post hoc for the largest relative gauge-RMSE reduction.
(c) Gauge RMSE on all eight hidden basins.}
  \label{fig:app-rsi-tidal}
\end{figure}

\paragraph{Task and source of difficulty.}

The agent must estimate a spatially varying seabed friction field from bathymetry, boundary forcing, and sparse observations of complex tidal elevation.
Some gauges contain large, unmarked observation errors.
Because the forward simulator smooths the effect of friction changes on tidal elevation, the inverse problem is ill-posed: fitting the observed gauges does not guarantee accurate predictions elsewhere.
The submitted solver must return a bounded $24\times24$ log-friction field.
Quality is measured by tidal-elevation root-mean-square error (RMSE) at separate gauges, rather than by direct agreement between the estimated and true friction fields.
Development uses six public basins, and final evaluation uses eight hidden cases.
Figure~\ref{fig:app-rsi-tidal} shows the research process, spatial error comparisons, and results on all eight hidden basins.


\begin{figure}[t]
  \centering
  \includegraphics[width=\linewidth]{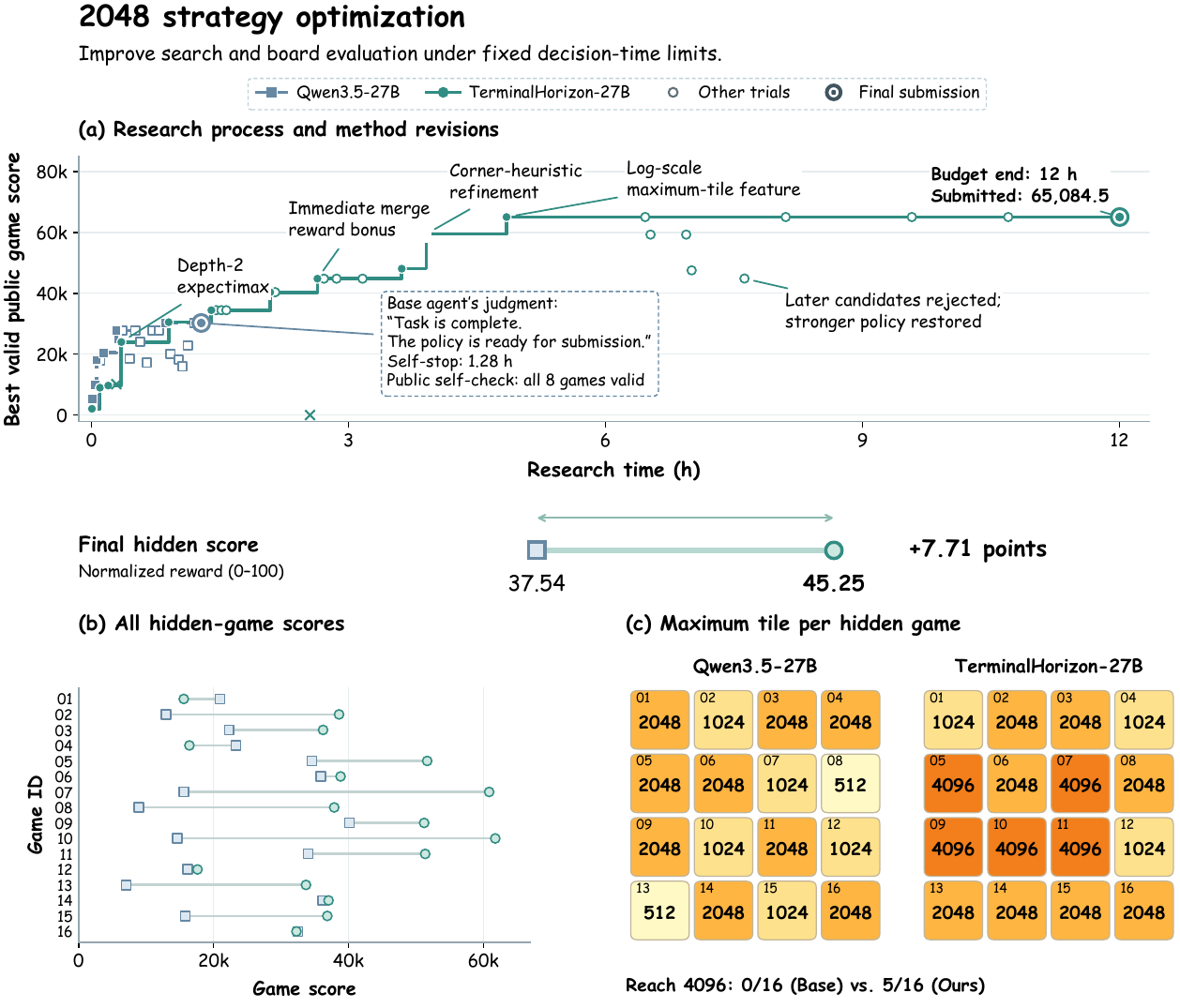}
\caption{2048 strategy optimization.
(a) Best valid mean score on eight public seeds. Crosses mark invalid
trials and rings mark submissions.
(b--c) Scores and maximum tiles for all 16 hidden games in seed order.
Each grid cell represents one game.}
  \label{fig:app-rsi-2048}
\end{figure}

\paragraph{Revising the model of the inverse problem.}

The base model recognizes the need for spatial regularization and outlier handling, adopting a $4\times4$ bilinear parameterization, coordinate descent, and iterative residual-based observation rejection.
After achieving a public mean RMSE of 0.023616, it voluntarily stops after 1.12 hours.
Before stopping, it judges the task complete because the solver has improved significantly over the initial implementation and is ready for evaluation.
TerminalHorizon-27B instead continues to compare spatial bases, robust losses, and regularization methods rather than treating the initial parameterization as fixed.
Its final method combines blockwise spatial coefficients with depth and log-depth features, penalizes differences between adjacent coefficients, and fits the complex-valued observations with a Huber loss.
It then removes the two gauges with the largest residuals and refits.
These choices address different sources of error, including limited spatial representation, unstable variation in the friction field, and corrupted observations.


\paragraph{Generalization beyond the fitted observations.}

The trajectory includes comparisons on reserved public cases and additional synthetic basins, allowing the agent to test whether a configuration's improvements extend beyond the cases used for tuning.
The final submission achieves a public mean RMSE of 0.011118, compared with 0.023616 for the base model.
On hidden cases, mean RMSE decreases from 0.021873 to 0.009386, a reduction of approximately 57.1\%, while the normalized score rises from 27.41 to 48.15.
This case shows how sustained research goes beyond additional optimization iterations by using experimental feedback to revise the friction-field representation and outlier handling, then checking whether those changes generalize to cases not used for tuning.


\subsection{2048 Strategy Optimization}
\label{app:rsi-2048}

\paragraph{Task and source of difficulty.}

The agent must improve a deterministic policy for the stochastic game 2048.
Given a board, the policy must select a legal move using only the Python standard library while meeting a per-move latency limit and a shared CPU time budget.
Development uses eight public game seeds, and the final policy is evaluated on 16 additional, non-overlapping seeds.
These requirements create competing objectives: deeper search may improve decisions but increases computational cost, while heuristics that work well on a few seeds may not generalize to other random tile sequences.
Figure~\ref{fig:app-rsi-2048} shows key search revisions and compares the final policies on each hidden game.


\paragraph{Search, evaluation, and selective revision.}

Both models improve on the weak initial implementation by introducing expectimax search and revising the board evaluation function.
The base model explores lookahead search that accounts for random tile generation, adaptive search depth, and corner preferences, then restores its best public-set policy and voluntarily stops after 1.28 hours.
It cites the improvement over the initial policy and the valid completion of all eight public games as grounds for submission, concluding that the task is complete and the policy is ready to submit.
TerminalHorizon-27B continues to introduce an immediate merge reward and selective memoization, while adjusting the relative influence of empty-cell count, tile values, smoothness, and monotonicity.
The development record also shows recovery from failed implementations: a version with unbounded cache growth fails to complete the full public evaluation, prompting a more selective caching design.
Later, changing the maximum-tile feature to a logarithmic form raises the mean score on the full public set from 59,509.0 to 65,084.5.


\paragraph{Improvement includes rejecting attractive candidates.}

Several later configurations perform well on smaller development subsets but regress on the full public test set.
Rather than replacing its best-so-far solution with the latest candidate, TerminalHorizon-27B reverts these changes and submits the stronger evaluated policy at the deadline.
The final public mean scores are 30,257.0 for the base model and 65,084.5 for TerminalHorizon-27B.
On hidden games, mean score increases from 23,183.75 to 38,626.25, while the normalized score rises from 37.54 to 45.25.
Both final policies achieve a valid-run fraction of 1.0 on the hidden set, so the difference mainly concerns policy quality rather than avoidance of execution failures.
This trajectory illustrates how sustained experimentation combines algorithmic improvement with careful selection: failed attempts provide evidence for retaining, revising, or discarding a candidate.


\subsection{Multi-View Point-Cloud Fusion}
\label{app:rsi-pointcloud}

\begin{figure}[t]
  \centering
  \includegraphics[width=\linewidth]{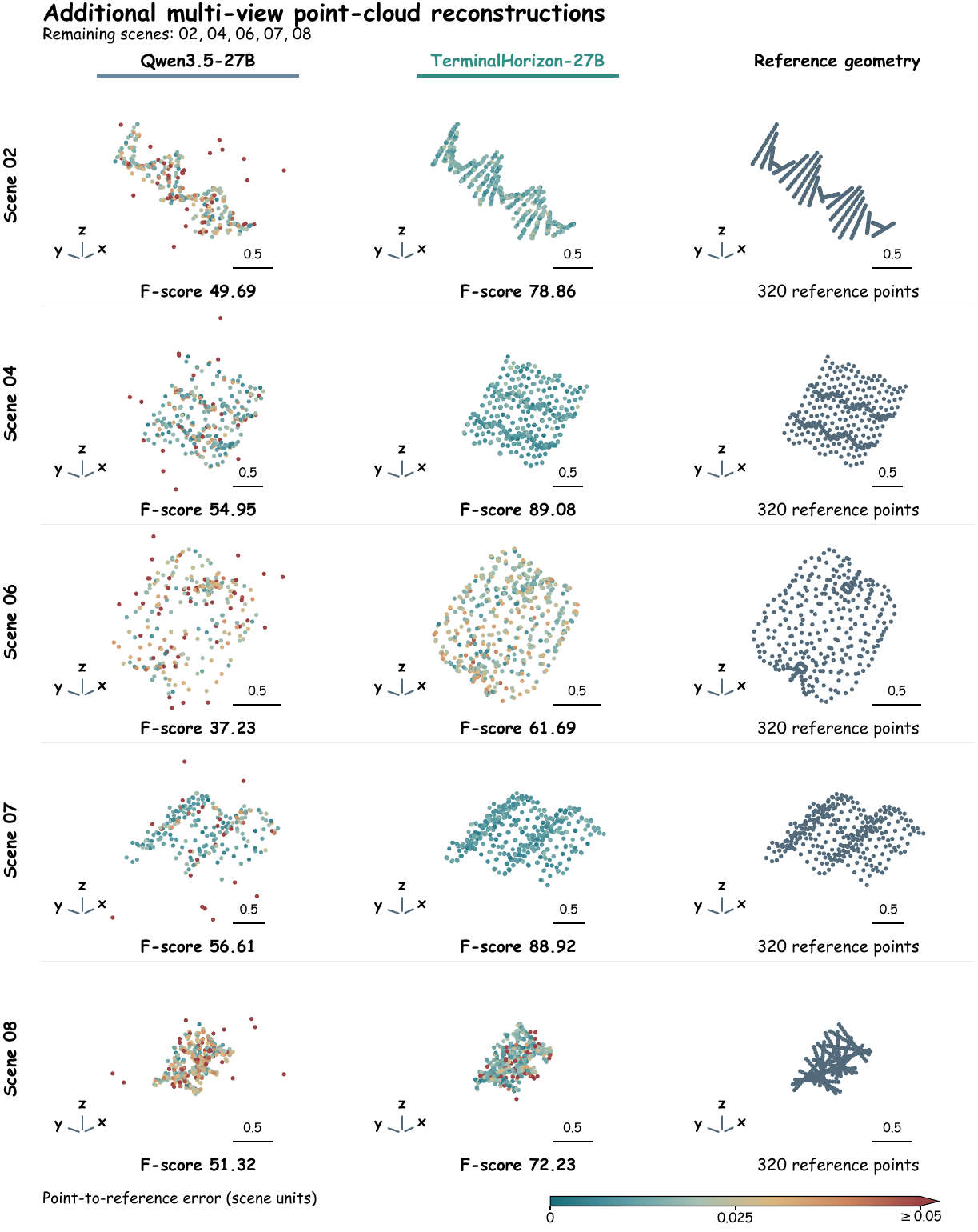}
    \caption{Additional point-cloud reconstructions
for five scenes beyond Figure~\ref{fig:long-horizon-applications}.
Each row uses the same camera, scale, and point size, without post-hoc
alignment, smoothing, or point removal.
Colors indicate point-to-reference distance, saturated at 0.05 scene units.
Scale bars represent 0.5 scene units, and F-scores are multiplied by 100.}
  \label{fig:app-rsi-pointcloud}
\end{figure}

\paragraph{Task and source of difficulty.}

The agent must reconstruct a complete scene by fusing noisy point clouds from partially overlapping views.
The first view defines the reference coordinate frame, while the remaining views contain unknown small rigid transformations; each observed point also has a confidence value.
The agent must align the views, suppress unreliable points, and preserve surface coverage within a fixed output-size limit.
The public self-check only verifies compliance with the output interface and provides neither ground-truth geometry nor reconstruction quality scores.
Final evaluation uses nine unseen scenes and averages symmetric precision--recall F-scores over three distance tolerances, penalizing both inaccurate points and missing surface regions.
Figure~\ref{fig:app-rsi-pointcloud} shows the five remaining scenes (02, 04, 06, 07, and 08), complementing scenes 01, 03, 05, and 09 in Figure~\ref{fig:long-horizon-applications}.
These visualizations compare final submissions rather than intermediate research versions.


\paragraph{Questioning what a proxy actually measures.}

After trying iterative closest point alignment, the base model settles on a conservative approach: it translates the other views toward the reference view's centroid and retains only points close to that view.
It obtains a high F-score on a self-designed proxy metric and voluntarily stops after 12.1 minutes.
Specifically, it describes the results as excellent based on a mean proxy F-score of 0.8994 and a proxy F-score of 1.0 at the 0.05 distance threshold for every case, concluding that the solver is ready to submit.
However, agreement with the reference observation is not equivalent to agreement with the complete scene.
Selecting points close to a single view can raise this proxy score while omitting surfaces visible only in other views.
TerminalHorizon-27B encounters a similar issue in early experiments.
A variant that selects points by nearest distance to the reference view achieves a proxy score of about 0.989, but the agent inspects which views contribute the selected points and questions whether the metric rewards reconstruction or merely copying the reference view.
It then introduces a leave-one-view-out diagnostic, identifies poor coverage of the excluded view, and revises its fusion strategy accordingly.


\paragraph{Aligning the method with geometric coverage.}

TerminalHorizon-27B's final solver filters nearest-point correspondences by a distance threshold, estimates rotation and translation through rigid alignment, and jointly refines each non-reference view against the other aligned views.
It performs two rounds of global refinement and selects high-confidence points from the merged cloud, rather than restricting added points to the area near the first view.
In the official hidden evaluation, the mean F-score across distance tolerances, expressed on a 0--100 scale, increases from 48.04 to 77.68; the separately calibrated normalized score rises from 21.93 to 48.96.
The key behavioral difference is the agent's willingness to revise not only the solver but also the experiments used to assess progress.
Interface validity and high proxy scores provide only intermediate evidence, while cross-view diagnostics help align the method with the actual reconstruction objective.


Together, these four cases illustrate complementary aspects of sustained autonomous research: jointly optimizing coupled objectives, revising scientific models, testing and reverting search heuristics, and identifying misleading proxy metrics.
They share a close connection between observations, subsequent decisions, and the implementation ultimately submitted for evaluation.
Longer interaction is valuable when it supports these connections, rather than serving as an objective in itself.


\section{Professional Domain Coverage}
\label{app:professional-domain-coverage}
\begin{figure}[t]
  \centering
  \includegraphics[width=\linewidth]{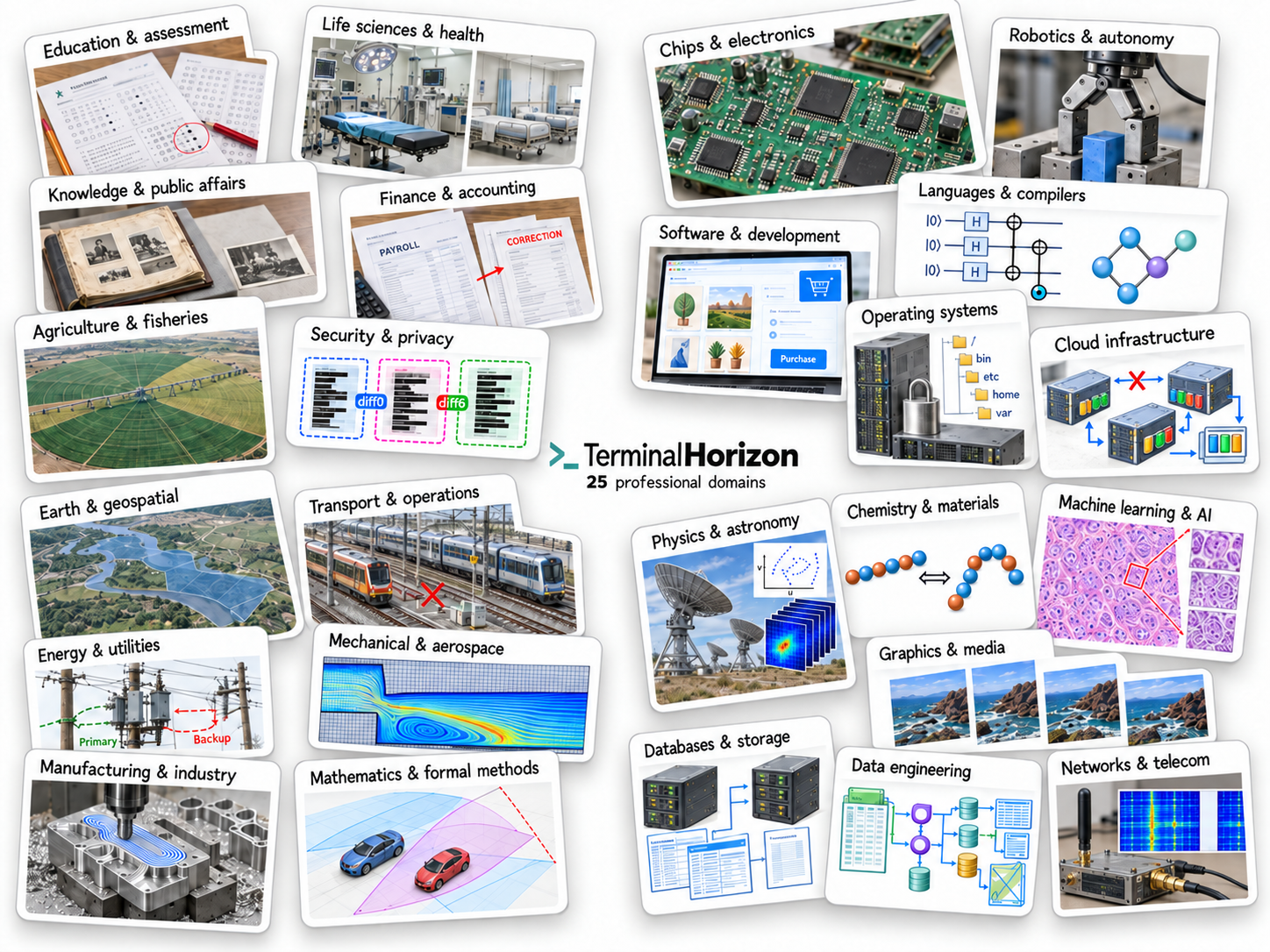}
  \caption{Overview of TerminalHorizon's 25 professional domains.}
  \label{fig:app-professional-domains}
\end{figure}

Our 1,535 tasks cover 25 primary professional domains.
As illustrated in Figure~\ref{fig:app-professional-domains}, this coverage extends beyond software and infrastructure to education, agriculture, healthcare, finance, and scientific research.


The terminal provides a shared interaction interface rather than defining the scope of the work.
Tasks such as educational assessment, payroll reconciliation, and radio-interferometric calibration bring different materials, constraints, and verifiable outputs into executable environments.
Domain coverage describes what agents work on, while the behavioral patterns in Appendix~\ref{app:agentic-behavioral-patterns} describe how they organize and advance that work.


\begin{table}[t]
  \centering
  \setlength{\belowcaptionskip}{6pt}
  \caption{Agentic behavioral patterns (1/2): exploration and diagnosis.}
  \label{tab:appendix-behavioral-patterns}
  \begingroup
  \fontsize{9}{11}\selectfont
  \setlength{\tabcolsep}{5pt}
  \renewcommand{\arraystretch}{1.08}
  \setlength{\heavyrulewidth}{0.7pt}
  \setlength{\lightrulewidth}{0.35pt}
  \arrayrulecolor{black!60}
  \begin{tabularx}{\linewidth}{
    >{\raggedright\arraybackslash}p{0.31\linewidth}
    >{\raggedright\arraybackslash}X}
    \toprule
    \rowcolor{nblue!6}
    \textbf{Pattern} & \textbf{Definition} \\
    \midrule
    \rowcolor{darker!7}
    \multicolumn{2}{l}{\strut\textbf{Information exploration and judgment}} \\
    \addlinespace[2pt]
    Problem-driven exploration
      & Using findings from earlier exploration to redirect subsequent searches toward different objects or scopes. \\
    Relational tracing
      & Following call, data-flow, identity, or provenance links to understand how objects relate or where to intervene. \\
    Multi-source evidence integration
      & Establishing correspondences across sources and combining their information to support a judgment. \\
    Candidate comparison and selection
      & Comparing alternatives against explicit criteria and using the comparison to select a candidate or defer selection. \\
    Tool and interface probing
      & Experimenting with unfamiliar tool or interface behavior and applying the resulting understanding in subsequent actions. \\
    \addlinespace[3pt]
    \rowcolor{darker!7}
    \multicolumn{2}{l}{\strut\textbf{Diagnosis and correction}} \\
    \addlinespace[2pt]
    Error-directed correction
      & Correcting an issue directly identified by an error or validation failure. \\
    Hypothesis-driven diagnosis
      & Testing a specific explanation or proposition and using the outcome to strengthen, weaken, or revise it. \\
    Reproduction-based diagnosis
      & Establishing and testing inputs or steps that reproduce an anomaly, then using them to investigate it further. \\
    Differential and partition-based localization
      & Comparing or partitioning versions, inputs, or components to narrow the suspected source of a problem. \\
    Runtime-guided localization
      & Using traces, profiles, breakpoints, or sampled internal behavior to narrow the source of a problem. \\
    Modification--recheck iteration
      & Modifying a visible problem and rechecking that same problem on the modified artifact, regardless of the outcome. \\
    \bottomrule
  \end{tabularx}
  \endgroup
\end{table}

\begin{table}[t]
  \centering
  \setlength{\belowcaptionskip}{6pt}
  \caption{Agentic behavioral patterns (2/2): construction, verification, and coordination.}
  \label{tab:appendix-behavioral-patterns-continued}
  \begingroup
  \fontsize{9}{11}\selectfont
  \setlength{\tabcolsep}{5pt}
  \renewcommand{\arraystretch}{1.08}
  \setlength{\heavyrulewidth}{0.7pt}
  \setlength{\lightrulewidth}{0.35pt}
  \arrayrulecolor{black!60}
  \begin{tabularx}{\linewidth}{
    >{\raggedright\arraybackslash}p{0.31\linewidth}
    >{\raggedright\arraybackslash}X}
    \toprule
    \rowcolor{nblue!6}
    \textbf{Pattern} & \textbf{Definition} \\
    \midrule
    \rowcolor{darker!7}
    \multicolumn{2}{l}{\strut\textbf{Construction and work progression}} \\
    \addlinespace[2pt]
    Subproblem decomposition and composition
      & Addressing distinct subproblems separately and combining their results to advance the main objective. \\
    Dependency-driven progression
      & Establishing prerequisites that subsequent work actually uses. \\
    Incremental construction and integration
      & Extending an existing partial capability or result and integrating additions into a more complete whole. \\
    Measurement--comparison--tuning
      & Comparing measurements before and after an intervention and using them to retain, adjust, or replace it toward an improvement goal. \\
    Item-wise processing and exception handling
      & Applying a common operation to a collection with identifiable item-level results, optionally routing exceptional items to separate handling. \\
    \addlinespace[3pt]
    \rowcolor{darker!7}
    \multicolumn{2}{l}{\strut\textbf{Verification and result establishment}} \\
    \addlinespace[2pt]
    Multi-check verification
      & Combining distinct, meaningful checks to support or qualify an outcome. \\
    Reference-based verification
      & Comparing an actual result against an explicit expected relation to determine agreement or discrepancy. \\
    Independent cross-verification
      & Checking the same claim through substantively different derivations, measurements, or implementations with distinct sources of error. \\
    Stability rechecking
      & Repeating and comparing runs under a fixed setup to assess variability, intermittency, or repeatability. \\
    Revalidation under changed conditions
      & Rechecking an existing claim after changing key inputs, environments, scales, operating conditions, or restart states. \\
    Requirement and delivery reconciliation
      & Revisiting task requirements, matching them to actual results or gaps, and using that assessment to deliver or continue work. \\
    \addlinespace[3pt]
    \rowcolor{darker!7}
    \multicolumn{2}{l}{\strut\textbf{Strategy and execution coordination}} \\
    \addlinespace[2pt]
    Evidence-driven strategy revision
      & Changing an established solution method or consequential arrangement in response to new information. \\
    Exploratory branching and backtracking
      & Trying a branch, then using feedback to undo its changes or return to an earlier decision state. \\
    Coupled-object coordination
      & Coordinating operations across objects linked by interfaces, identities, timing, resources, or state, and checking their joint effects. \\
    Asynchronous job management
      & Using identifiable background jobs' states to guide subsequent management, collect results, or establish unfinished status. \\
    Sequential decision-making in dynamic environments
      & Selecting subsequent actions in response to evolving task-world conditions, including consequences of earlier actions. \\
    \bottomrule
  \end{tabularx}
  \endgroup
\end{table}

\section{Agentic Behavioral Patterns}
\label{app:agentic-behavioral-patterns}

\subsection{Definitions and Taxonomy}
\label{app:pattern-definitions}
We define an \emph{agentic behavioral pattern} as the structure connecting actions, observations, and subsequent decisions around a specific goal during problem solving.
Unlike classifications by task domain or tool, behavioral patterns describe how agents gather information, address problems, construct solutions, and verify their correctness, rather than characterizing individual commands or isolated operations.


We group 27 behavioral patterns into five categories: information exploration and judgment, diagnosis and error correction, construction and work progress, verification and assurance, and strategy and execution coordination.
Tables~\ref{tab:appendix-behavioral-patterns} and~\ref{tab:appendix-behavioral-patterns-continued} define each pattern.
These categories describe different aspects of problem-solving behavior rather than mutually exclusive task types or fixed execution stages: a trajectory may contain multiple patterns, and tasks from different domains may exhibit the same patterns.


\begin{figure}[hp]
  \centering
  \includegraphics[width=\linewidth]{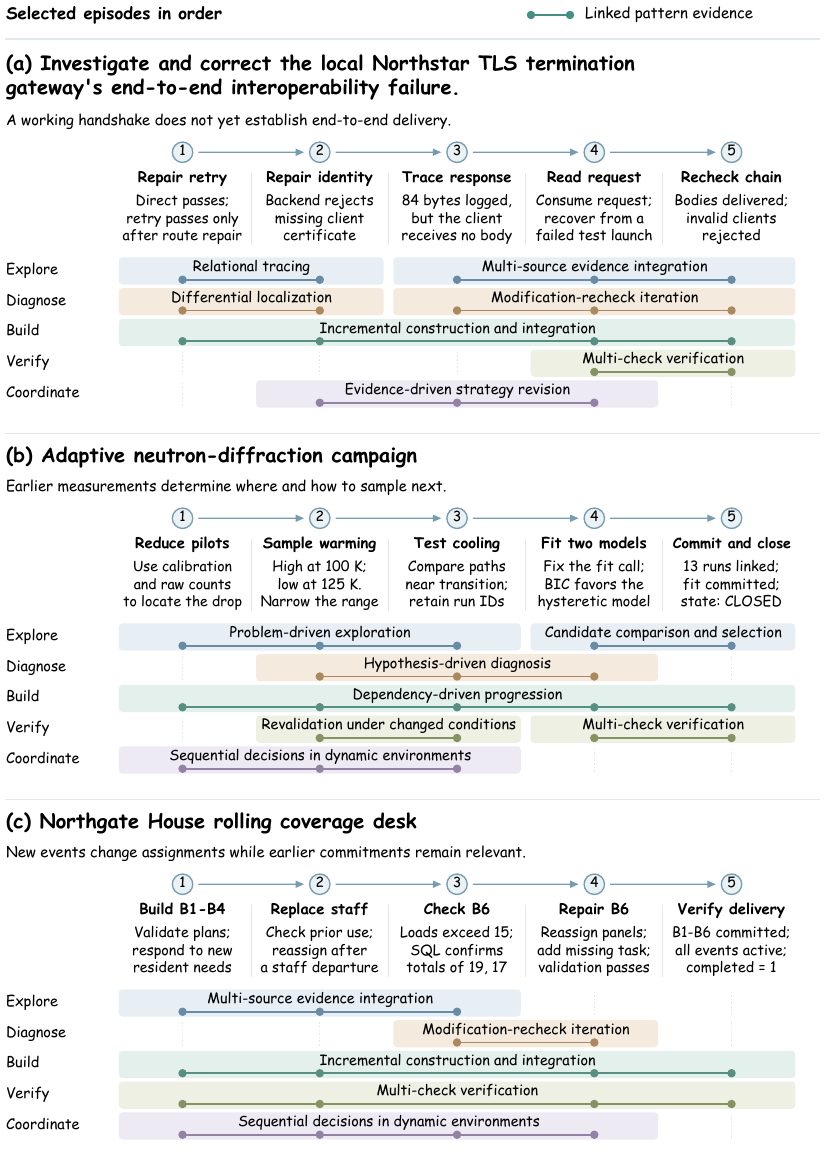}
  \caption{Behavioral-pattern organization in three trajectories.
  Columns show selected episodes in order. Dots mark supporting episodes, linked by pattern within each family.
  Horizontal spacing indicates order, not duration.}
  \label{fig:app-pattern-organization}
\end{figure}

\subsection{Organization of Behavioral Patterns within Trajectories}
\label{app:pattern-organization}

To examine how agentic behavioral patterns connect within a trajectory, we select three annotated cases with traceable links between observations, decisions, and subsequent work.
Figure~\ref{fig:app-pattern-organization} presents a TLS gateway repair, a synthetic neutron-diffraction study, and an event-driven staffing task.
The figure links selected episodes to observed patterns from the five categories above.
These are qualitative examples of behavioral organization, rather than estimates of pattern frequency or duration.


\paragraph{TLS gateway repair: diagnosis returns as implementation progresses.}
The agent first distinguishes a working direct handshake from a failing retry path and corrects the route handling.
Backend logs then expose a missing client identity, which requires a separate repair.
Even after that repair, the downstream client receives no response body despite 84 bytes being recorded in the gateway log, a discrepancy that redirects the investigation toward request consumption and connection closure.
After modifying the request-reading path and recovering from a failed test launch, the agent reruns the end-to-end checks, including invalid-client rejection.
Relational tracing, diagnosis, and modification--recheck iteration thus support incremental construction rather than occurring as isolated stages.
The final evidence is a passing local regression suite, not an exhaustive check of every TLS requirement.


\paragraph{Neutron diffraction: observations guide sampling and model selection.}
The agent combines instrument calibration with raw pilot measurements to locate a drop in the signal.
New warming measurements narrow the temperature range, while cooling measurements provide a comparison near the transition.
These observations support a test of thermal-path dependence and a BIC comparison between continuous and hysteretic models.
The selected fit is committed with all 13 run identifiers before the campaign is closed.
Problem-driven exploration, hypothesis-driven diagnosis, and dependency-driven progression are connected through the measured data.
The native interface records the fit and closure, but the trajectory does not expose the independent verifier's refit results.


\paragraph{Care staffing: changing conditions connect planning, correction, and delivery.}
In the synthetic staffing environment, six consecutive windows require updates for resident transfers, changing care needs, admissions, and staff availability.
Each committed window contributes to the cumulative schedule and constrains later assignments.
For the final window, validation reports two panel loads of 19 and 17 against a capacity of 15, along with a missing licensed task.
A separate SQL calculation confirms the loads before the agent redistributes residents, adds the task, and revalidates the plan.
The final status records all six windows as committed.
This case connects dynamic decision-making and incremental construction with a local correction loop and final delivery checks.
The SQL calculation and validator use different checking paths but share the same database inputs.


Across these cases, a single episode can support several patterns, and one pattern can connect work across multiple episodes.
The resulting experience lies in how evidence is used, partial results are extended, and checks determine the next action, rather than in a fixed sequence of behavioral stages.


\section{Trajectory Filtering Criteria}
\label{app:trajectory_filtering}

Final task success alone does not determine a trajectory's value for
supervised learning. Successful trajectories may rely on answer leakage
or circumvent task constraints, whereas unsuccessful ones may still
demonstrate useful diagnosis, implementation, and error recovery.
We therefore adopt process-oriented, trajectory-level filtering based on
trajectory validity, environment-grounded progress, and multi-turn training
value to assess whether a teacher trajectory provides a suitable
interaction demonstration for the student model.
We use \texttt{gpt-5.6-luna} with reasoning effort set to \texttt{high}
as the reviewer. It reads the complete conversation without receiving
the final reward or a success/failure label as additional input,
and must ground its judgments in the actions and observations
recorded in the trajectory.


\paragraph{Trajectory validity.}

A valid trajectory should maintain coherent context and traceable
action--observation correspondences, allowing decisions to be connected
to their execution outcomes.
We exclude trajectories whose core problem-solving process cannot be
reliably interpreted because of severe truncation, persistent protocol
corruption, or an incorrect execution environment.
We also exclude trajectories that rely on leakage of private answers,
unauthorized manipulation of internal state, or circumvention of
prescribed interfaces.
Local interaction errors do not, by themselves, invalidate a trajectory:
if the model recognizes a parsing or command error, restores valid
interaction, and continues making progress, the sequence can still
provide a useful demonstration of error correction.
This criterion distinguishes recoverable mistakes from substantive
defects that compromise trajectory integrity or violate task constraints.


\begin{table}[t]
    \centering
    \setlength{\belowcaptionskip}{6pt}
    \caption{Benchmark text used for decontamination. Paths are relative to each benchmark distribution.}
    \label{tab:decontamination_inputs}
    \begingroup
    \normalfont
    \fontsize{9}{11}\selectfont
    \renewcommand{\tabularxcolumn}[1]{m{#1}}
    \setlength{\tabcolsep}{6pt}
    \renewcommand{\arraystretch}{1.18}
    \setlength{\heavyrulewidth}{0.7pt}
    \setlength{\lightrulewidth}{0.35pt}
    \arrayrulecolor{black!60}
    \begin{tabularx}{\linewidth}{
        >{\raggedright\arraybackslash}m{96pt}
        >{\raggedright\arraybackslash}m{139pt}
        >{\raggedright\arraybackslash}X
    }
        \toprule
        \rowcolor{nblue!6}
        \textbf{Benchmark} & \textbf{Source file(s)} & \textbf{Compared content} \\
        \midrule

        Terminal-Bench\newline 1.0 / 2.0 / 2.1
        & instruction.md
        & Complete task instruction \\

        \rowcolor{black!2.5}
        AutomationBench
        & automationbench/domains/\newline
          \textless{}domain\textgreater{}/tasks.py
        & User messages in the generated task prompt \\

        VitaBench\newline (EN cross-domain)
        & cross\_domain/tasks\_en.json
        & instructions, user\_scenario, and nonempty message\_history \\

        \rowcolor{black!2.5}
        $\tau^2$-bench Airline
        & airline/tasks.json
        & Complete user\_scenario \\

        $\tau^2$-bench Retail
        & retail/tasks.json
        & Complete user\_scenario \\

        \rowcolor{black!2.5}
        MATH-500
        & test.jsonl
        & problem \\

        GSM8K
        & test-00000-of-00001.parquet
        & question \\

        \rowcolor{black!2.5}
        AIME 2024
        & test-00000-of-00001.parquet
        & problem \\

        AIME 2025
        & test.jsonl
        & problem \\

        \rowcolor{black!2.5}
        KOR-Bench (Cipher)
        & rule.jsonl; sample.jsonl
        & Rule title, rule content, and question \\

        \bottomrule
    \end{tabularx}
    \endgroup
\end{table}

\paragraph{Environment-grounded progress.}

High-quality trajectories should demonstrate that actions are informed
by evidence from the environment.
We examine whether the model uses file contents, logs, command outputs,
error messages, and test results to guide its decisions and, when
necessary, revise its hypotheses or plans.
Meaningful progress need not immediately produce a correct result:
ruling out an infeasible approach, identifying a root cause, or
uncovering a new constraint can also provide valuable learning signals.
Conversely, we exclude trajectories dominated by persistent disregard
for feedback, ungrounded trial and error, repeated ineffective actions,
or unsupported adherence to incorrect conclusions.
The assessment concerns the reliability of the overall action sequence,
rather than requiring every attempt to be correct.


\paragraph{Multi-turn training value.}

We further assess whether a trajectory demonstrates substantively
connected, multi-turn terminal work, including evidence gathering,
implementation, verification, hypothesis revision, and error recovery.
Its value lies in how interactions inform subsequent decisions,
not in trajectory length, command count, or output volume.
Waiting, repeated polling, and mechanical loops receive no additional
credit merely for increasing the number of turns.
Infrastructure operations used solely to connect to or manage the
execution environment likewise do not count toward problem-solving depth.
Rather than imposing a fixed turn-count threshold, we assess whether
the trajectory provides a reliable, nonredundant interaction process
worth learning from.


\paragraph{Overall decision.}

We retain trajectories that satisfy these criteria even if they do not
complete the task or end at an unresolved obstacle.
In particular, feedback-driven diagnosis, revision, and recovery
following a failure should not be discarded simply because they
contain mistakes. Local errors are likewise insufficient grounds for
rejection when they do not compromise the main action sequence.
We instead reject a trajectory when its core solution relies on
prohibited shortcuts, incorrect assumptions persist throughout the
main problem-solving process, or interactions remain unproductive
and fail to provide reliable learning signals.
These process-oriented criteria preserve exploration and error
correction in realistic problem solving while excluding systematic
errors and ineffective interactions from supervised demonstrations.




\suppressfloats[t]

\section{Detailed Benchmark Decontamination}
\label{app:benchmark_decontamination}

Direct reuse or close paraphrasing of evaluation problems in training tasks can make it difficult to distinguish capability gains from memorization.
To assess this risk, we compare all 1,535 training tasks with the evaluation benchmarks at the task level, combining exact-span matching, near-duplicate retrieval, and semantic assessment to identify potential task overlap.

%
\subsection{Comparison Corpora}
\label{app:decontamination-corpora}

%
The comparison corpora cover 3,251 evaluation instances across 12 benchmark groups: Terminal-Bench 1.0, 2.0, and 2.1; AutomationBench; the English Cross-domain tasks in VitaBench; $\tau^2$-bench Airline and Retail; MATH-500; GSM8K; AIME 2024; AIME 2025; and KOR-Bench Cipher.
Each training task is represented by its full \texttt{instruction.md}, while each evaluation instance is represented by its problem statement or user request.
Table~\ref{tab:decontamination_inputs} lists the sources and fields used.

\begin{table}[t]
    \centering
    \setlength{\belowcaptionskip}{6pt}
    \caption{Decontamination against 1,535 training tasks.
        $N$: benchmark instances, $E_n$/$M_8$: exact/math-aware overlap pairs,
        Contam./Reviewed: contaminated/reviewed pairs.}
    \label{tab:decontamination_results}
    \begingroup
    \fontsize{9}{11}\selectfont
    \setlength{\tabcolsep}{4pt}
    \renewcommand{\arraystretch}{1.18}
    \setlength{\heavyrulewidth}{0.7pt}
    \setlength{\lightrulewidth}{0.35pt}
    \arrayrulecolor{black!60}
    \renewcommand{\tabularxcolumn}[1]{m{#1}}
    \begin{tabularx}{\linewidth}{
        >{\raggedright\arraybackslash}X
        r c r r r r c
    }
        \toprule
        \rowcolor{nblue!6}
        \raisebox{5pt}{\textbf{Benchmark}}
        & \raisebox{5pt}{$N$}
        & \shortstack{\textbf{Exact}\\$E_{13}/E_{14}$}
        & \raisebox{5pt}{$M_8$}
        & \shortstack{\textbf{Max}\\$J_5$}
        & \shortstack{\textbf{Max}\\$S_{\mathrm{TF}}$}
        & \raisebox{5pt}{\textbf{JSD}}
        & \shortstack{\textbf{Contam./}\\\textbf{Reviewed}} \\
        \midrule

        Terminal-Bench 1.0
        & 80 & 0/0 & 8
        & 0.00175 & 0.07467 & 0.4334 & 0/19 \\

        \rowcolor{black!2.5}
        Terminal-Bench 2.0
        & 89 & 0/0 & 83
        & 0.00289 & 0.08344 & 0.3858 & 0/19 \\

        Terminal-Bench 2.1
        & 89 & 0/0 & 83
        & 0.00289 & 0.08345 & 0.3848 & 0/19 \\

        \rowcolor{black!2.5}
        AutomationBench
        & 600 & 0/0 & 0
        & 0.00196 & 0.07050 & 0.4447 & 0/14 \\

        VitaBench\newline (EN cross-domain)
        & 100 & 0/0 & 0
        & 0.00096 & 0.02761 & 0.5907 & 0/12 \\

        \rowcolor{black!2.5}
        $\tau^2$-bench Airline
        & 50 & 0/0 & 0
        & 0.00000 & 0.03764 & 0.6179 & 0/10 \\

        $\tau^2$-bench Retail
        & 114 & 0/0 & 0
        & 0.00177 & 0.03673 & 0.6068 & 0/13 \\

        \rowcolor{black!2.5}
        MATH-500
        & 500 & 0/0 & 15
        & 0.00602 & 0.07990 & 0.5914 & 0/18 \\

        GSM8K
        & 1,319 & 0/0 & 0
        & 0.00252 & 0.07342 & 0.5486 & 0/15 \\

        \rowcolor{black!2.5}
        AIME 2024
        & 30 & 0/0 & 0
        & 0.00179 & 0.06560 & 0.6280 & 0/13 \\

        AIME 2025
        & 30 & 0/0 & 6
        & 0.00381 & 0.06169 & 0.6547 & 0/18 \\

        \rowcolor{black!2.5}
        KOR-Bench (Cipher)
        & 250 & 0/0 & 210
        & 0.00360 & 0.05116 & 0.5878 & 0/20 \\

        \midrule
        \rowcolor{darker!7}
        \textbf{Total}
        & 3,251 & 0/0 & 405
        & -- & -- & -- & 0/190 \\
        \bottomrule
    \end{tabularx}
    \endgroup
\end{table}

%
The comparison texts retain the task objectives and necessary context.
For KOR-Bench Cipher, we combine the cipher rules with the question.
For VitaBench and $\tau^2$-bench, we use the full user-request scenario, preserving its goals, known information, and constraints, including requirements that may be disclosed gradually during interaction.
No text is summarized or truncated, and reference answers, private scoring criteria, and expected execution outcomes are excluded from the comparison.


\subsection{Text Overlap and Near-Duplicate Retrieval}
\label{app:decontamination-retrieval}

%
We apply Unicode NFKC normalization and lowercasing to the task descriptions, then tokenize them into contiguous sequences of Unicode letters, digits, and underscores.
Exact matching covers both the full normalized text and contiguous 13-token and 14-token windows with a stride of 1.
Let $G_n(x)$ denote the set of contiguous $n$-grams in task description $x$.
Exact-span overlap between a training description $x$ and an evaluation description $y$ is defined as
\begin{equation}
    E_n(x,y)
    =
    \mathbb{I}\!\left[G_n(x)\cap G_n(y)\neq\varnothing\right],
    \qquad n\in\{13,14\}.
\end{equation}

%
To retrieve near-duplicate tasks that do not share long verbatim spans, we also compute the Jaccard similarity between their 5-gram sets,
\begin{equation}
    J_5(x,y)
    =
    \frac{|G_5(x)\cap G_5(y)|}
         {|G_5(x)\cup G_5(y)|},
\end{equation}
and the cosine similarity $S_{\mathrm{TF}}(x,y)$ between their unigram/bigram TF--IDF representations.
The former captures local phrase reuse, while the latter captures lexical similarity under changes in word order.
TF--IDF is fitted on the combined corpus of training tasks and the benchmark being compared, using logarithmic term frequency and smoothed inverse document frequency.
For mathematical and symbolic expressions, we additionally retrieve contiguous 8-gram overlaps using a tokenizer that preserves \LaTeX{} commands, numbers, and operators.
Because these short spans may represent common formulas or interfaces, they supplement candidate retrieval rather than directly establish contamination.
We also report base-2 unigram Jensen--Shannon divergence (JSD) to characterize differences in word distributions between the training corpus and each benchmark; this metric is not used to determine contamination.

%
\subsection{Semantic Assessment of Candidate Pairs}
\label{app:decontamination-semantic}

%
Lexical similarity does not necessarily indicate task reuse, and a paraphrase of the same problem may not share long text spans.
We therefore combine multiple retrieval methods to select candidate pairs and assess whether they are substantive near-duplicates based on the complete task descriptions.
For each benchmark group, the candidate set includes all full-text and exact 13/14-gram matches, the top 10 pairs by TF--IDF similarity, the top 5 by 5-gram Jaccard similarity, and the top 5 by the number of shared math-aware 8-grams.
Each retrieval method contributes only pairs with a nonzero score, and the combined set is deduplicated before semantic review.

%
Semantic assessment is performed by \texttt{gpt-5.6-luna} with reasoning effort set to \texttt{high}.
The reviewer reads both task descriptions in full and compares their core objectives, initial conditions, key objects or data, distinctive combinations of constraints, required transformations, and completion criteria.
A pair is classified as contaminated when one task is a direct copy or minor paraphrase of the other, or when renamed entities leave the distinctive instance structure and solution requirements unchanged.
Shared tools, domain knowledge, terminology, or common work patterns alone do not constitute contamination.
Each pair is labeled as non-contaminated, contaminated, or uncertain, with supporting comparisons drawn from the task descriptions.

%
\subsection{Decontamination Results}
\label{app:decontamination-findings}

%
Table~\ref{tab:decontamination_results} summarizes text similarity and semantic assessment results for each benchmark.
None of the 12 benchmark groups contains an exact match to normalized full text or a contiguous 13/14-token window.
The maximum 5-gram Jaccard similarity is 0.006024, and the maximum TF--IDF cosine similarity is 0.083445.
Math-aware 8-gram retrieval identifies 405 pairs with short-span overlap, which are used for candidate selection alongside lexical nearest neighbors.
After combining and deduplicating the retrieval results, 190 candidate pairs undergo semantic review; none is classified as contaminated or uncertain.
These comparisons of task descriptions and semantic assessments find no evidence of evaluation-task reuse in the training tasks, so all 1,535 training tasks are retained.


\section{Case Studies of Challenge-Guided Reconstruction}
\label{app:challenge-guided-reconstruction}

We examine three cases of numerical calibration, distributed-service recovery, and household benefits administration in which independent challenges exposed work lost during environment construction.
Each traces the initial execution, the resulting reconstruction, and the work observed in a subsequent independent challenge.


\begingroup
\tcbset{
  reconstructioncase/.style={
    enhanced jigsaw,breakable,notitle after break,
    width=\linewidth,arc=2pt,boxrule=0.45pt,
    colback=white,colframe=darker!28!white,
    colbacktitle=white,coltitle=black,coltext=black,
    fonttitle=\rmfamily\bfseries\fontsize{10}{12}\selectfont,
    fontupper=\rmfamily\fontsize{9}{11}\selectfont,
    titlerule=0pt,lefttitle=8pt,righttitle=8pt,
    toptitle=7pt,bottomtitle=5pt,
    boxsep=0pt,left=6pt,right=6pt,top=0pt,bottom=6pt,
    before skip=9pt,after skip=8pt,
    pad at break*=6pt,lines before break=6
  },
  reconstructionstep/.style={
    enhanced,frame hidden,boxrule=0pt,arc=0pt,
    boxsep=0pt,left=8pt,right=8pt,top=5pt,bottom=5pt,
    before skip=3pt,after skip=3pt,
    coltext=black,
    fontupper=\rmfamily\fontsize{9}{11}\selectfont,
    before upper={\setlength{\parindent}{0pt}\setlength{\parskip}{2pt}
      \raggedright\emergencystretch=1em}
  }
}
\newcommand{\reconstructionstep}[3]{%
  \begin{tcolorbox}[reconstructionstep,colback=#1!4!white,
    borderline west={1.7pt}{0pt}{#1!55!white}]
    \textbf{#2}\par\nobreak
    #3
  \end{tcolorbox}%
}

\subsection{Radio-Interferometric Calibration}
\label{app:reconstruction-radio-calibration}

The agent must calibrate a synthetic Northstar-10 radio observation stored as a native MeasurementSet while preserving the raw measurements and metadata.
Calibration is evaluated on holdout and target data excluded from fitting.
The environment uses casacore, NumPy, and SciPy for data access and numerical computation.


\begin{tcolorbox}[reconstructioncase,
  title={Case 1\enspace\textbar\enspace Restoring numerical calibration work}]

\reconstructionstep{nblue}{01\quad Initial construction}{%
The public \texttt{mscal} tool included \texttt{solve-bandpass} and
\texttt{solve-gain}, which performed numerical fitting and wrote native
calibration tables, so core implementation work was already packaged into the environment.}

\reconstructionstep{nblue}{02\quad Independent challenge: successful, but compressed}{%
After identifying the fields, the challenger invoked the bundled solvers,
applied the calibration, and checked residuals. It adjusted the fitted field selection
after a failed check, but did not need to implement the numerical calibration.}

\reconstructionstep{darker}{03\quad Reconstruction: leave the fit to the agent}{%
Reconstruction removed the fitting routines but retained inspection, calibration
application, and residual feedback. It introduced distinct time-varying responses
for the two spectral windows and made the required solution axes and table conventions public.}

\reconstructionstep{darker}{04\quad Independent re-challenge: feedback changes the model}{%
The fresh challenger inspected the data axes and public interfaces, implemented
a fitting workflow, and revised its spectral-window-dependent parameters after
observing high residuals. It wrote and applied native calibration tables,
checked that the raw data remained unchanged, and evaluated holdout and target data without fitting them.
\par
Final weighted RMS was $0.896$ on the fitted calibrators, $1.784$ on holdout,
and $1.944$ on target, all below the public threshold of $2.8$.
The independent run passed the verifier. These scores describe the reconstructed
task, not a comparison between task versions.}
\end{tcolorbox}

\paragraph{What reconstruction restores.}
The initial challenger also passed, showing that task success alone does not reveal work completed in advance.
Reconstruction moved numerical modeling and implementation from the supplied tool to the agent while retaining the interfaces and feedback needed to complete the task.
The resulting trajectory connects model construction, residual-driven revision, and independent validation.


\subsection{RabbitMQ Recovery Across Broker and Application State}
\label{app:reconstruction-rabbitmq}

The agent must recover a three-node RabbitMQ service from a network partition while preserving confirmed messages, resolving failed or uncertain publications, and preventing duplicate business effects.
The synthetic incident runs on real RabbitMQ brokers and AMQP clients, with firewall rules and persistent application state.
Five reconstruction rounds and six independent challenges turned application recovery from a possible branch into work required by the running environment.


\begin{tcolorbox}[reconstructioncase,
  title={Case 2\enspace\textbar\enspace From broker recovery to application recovery}]

\reconstructionstep{nblue}{01\quad Initial challenge: recovery is largely prebuilt}{%
The challenger used the supplied \texttt{partitionctl heal}, restarted one broker,
and let the clients finish. Automatic retry and idempotent consumption handled
much of the application recovery, leaving a short operational path.}

\reconstructionstep{darker}{02\quad Rounds 1--2: reconstruct the live recovery state}{%
The redesigns removed the healing command and introduced three queues with
different missing voting replicas. Missing members and network faults were selected
from live broker state rather than a fixed repair map. Unresolved publications
were kept in a durable outbox, and recovery had to preserve unrelated firewall policy.}

\reconstructionstep{darker}{03\quad Round 3: restore dependencies within application work}{%
The workload changed from independent messages to 27 workflows of eight ordered
operations. Publishers retained event identities for reconciliation and resumption;
consumers tracked predecessor completion and prevented duplicate business effects.
The intended recovery now depended on both broker health and the completion of earlier workflow steps.}

\reconstructionstep{nblue}{04\quad Repeated challenges: the new work still does not occur}{%
Intermediate challenges still finished without the intended reconciliation work.
One run encountered no uncertain publications; in a later run, the client could
reach its local queue leader despite the partition. The recovery branch existed
but was not needed in either execution.}

\reconstructionstep{darker}{05\quad Rounds 4--5: connect the fault to its application consequence}{%
The redesigns moved fault installation before client startup and connected the
orders client to a runtime-selected follower whose leader was across the partition.
The actual AMQP path now exposed the application failure that earlier versions
could avoid.}

\reconstructionstep{darker}{06\quad Final independent challenge: recover the missing predecessor}{%
After restoring connectivity and queue voters, the challenger found 71 confirmed
orders rows and one \texttt{nacked} first-step event, with downstream consumer
errors. It retrieved the stored message body and original event ID, then
republished with AMQP publisher confirms. All 216 outbox rows became confirmed,
every workflow reached step 8, and no duplicate business effects were observed.
The independent run passed.}
\end{tcolorbox}

\paragraph{What reconstruction restores.}
Reconstruction linked network and replica repair to unfinished application work.
Repeated challenges showed that a recovery branch adds no workload unless the running environment actually triggers it.
The final run had to recover a specific missing event before dependent workflow steps could finish.


\subsection{Family Enrollment Changes and Payroll Reconciliation}
\label{app:reconstruction-benefits}

This case concerns a household coverage change following the loss of other insurance.
The agent must assess dependent eligibility and keep benefits enrollment, payroll deductions, and insurer records consistent.
The household, rules, and service are synthetic, modeling an employer's benefits workflow through stateful public interfaces.
Four reconstruction rounds and five independent challenges developed a simple record-update task into evidence-based review and coordinated updates across systems.


\begin{tcolorbox}[reconstructioncase,
  title={Case 3\enspace\textbar\enspace From enrollment edits to evidence-based reconciliation}]

\reconstructionstep{nblue}{01\quad Initial challenge: a fixed update sequence}{%
After inspecting the case, the challenger amended three plans, denied one
dependent, and synchronized payroll and insurer records. Most extra interactions
came from discovering command syntax, not from analyzing evidence or revising decisions.}

\reconstructionstep{darker}{02\quad Rounds 1--2: connect coverage dates to downstream records}{%
Previously posted deductions now required retroactive payroll adjustments.
A dependent's age limit split coverage into consecutive periods, changing
pay-period calculations and insurer payloads. Legacy insurer plan codes also
had to be updated before the new records could be accepted.}

\reconstructionstep{darker}{03\quad Round 3: revise an already accepted case}{%
The starting state now contained committed enrollment, applied payroll, and an
accepted insurer transaction, followed by a corrected dependent document.
The agent had to reopen the case, create a replacement version, and update both
downstream systems without erasing the earlier records.}

\reconstructionstep{nblue}{04\quad Re-challenge: the evidence decision is still supplied}{%
The corrected document was already marked verified, the affected person was
obvious, and each downstream issue had one clear remedy. The challenger could
still follow a fixed sequence once it had read the case. Versioned updates alone
had not restored the intended evidence analysis.}

\reconstructionstep{darker}{05\quad Round 4: make review outcomes change the decision}{%
Reconstruction introduced conflicting document claims and their source records.
One dependent had both a mismatching document and a corroborating alternative;
another had an authentic document whose loss date failed the event-date rule.
Public review and validation returned feedback that determined which people
could be included in the revised coverage.}

\reconstructionstep{darker}{06\quad Final independent challenge: revise and propagate the decision}{%
The challenger rejected mismatching evidence, found the corroborating source,
and revised its provisional enrollment after validation exposed the ineligible
loss date. It approved D-4, denied D-3 and D-5, and preserved the age-bounded
coverage for D-2. It then committed version 6, applied a \$312 payroll adjustment,
and obtained insurer acceptance while retaining version 5. The independent run passed.}
\end{tcolorbox}

\paragraph{What reconstruction restores.}
The key change is that document review now determines the next action rather than confirming a decision already supplied by the environment.
An authentic document can still fail the eligibility rule, and an unsupported claim requires another source.
Once the household decision changes, its effective dates and source version must remain consistent across enrollment, payroll, and insurer acceptance.
The final trajectory therefore connects evidence comparison, feedback-driven revision, and verified updates across systems.


\endgroup

\section{Qualitative Analysis of Progressive Task Evolution}
\label{app:progressive-task-evolution}

We examine three evolution lineages from two original tasks: two branches of photo-library reconciliation and one hotel event-planning lineage.
For each generation, we examine what changes from its immediate parent, why the previous approach is no longer sufficient, and which additional operations and checks are needed.
The photo branches show how one task can develop in different directions, while the hotel lineage shows how new constraints change earlier planning decisions.


\subsection{Two Branches from the Same Photo-Library Task}
\label{app:evolution-photo}

The shared parent asks the agent to reconstruct a curated PhotoPrism library from media exports, catalogs, and sidecars.
The work includes preserving distinct media files, organizing related representations, restoring curation and albums, and checking the native library after a complete rescan.
Table~\ref{tab:app-photo-evolution} compares two three-generation branches that start from identical parent packages.


\begin{table}[!htb]
  \centering
  \setlength{\belowcaptionskip}{6pt}
  \caption{Two photo-library evolution branches from the same parent task. Numbers indicate successive generations within each branch.}
  \label{tab:app-photo-evolution}
  \begingroup
  \normalfont
  \fontsize{9}{11}\selectfont
  \setlength{\tabcolsep}{5pt}
  \renewcommand{\arraystretch}{1.12}
  \setlength{\extrarowheight}{2pt}
  \renewcommand{\tabularxcolumn}[1]{m{#1}}
  \setlength{\heavyrulewidth}{0.7pt}
  \setlength{\lightrulewidth}{0.35pt}
  \arrayrulecolor{black!60}
  \begin{tabularx}{\linewidth}{
    >{\centering\arraybackslash\bfseries}m{32pt}
    >{\raggedright\arraybackslash}X
    >{\raggedright\arraybackslash}X}
    \toprule
    \rowcolor{nblue!6}
    \textbf{Version} & \textbf{Change from the parent} & \textbf{Additional work} \\
    \midrule

    \rowcolor{darker!7}
    \multicolumn{3}{l}{\strut\textbf{Branch A: reconciling changes in library state}} \\
    \addlinespace[3pt]
    A1
      & A one-way import becomes a second synchronization with independent edits on both sides. A shared photo's rating changes from 5 at the checkpoint to 4 at the source and 3 at the destination.
      & Compare checkpoint, source, and destination values. Preserve legitimate edits from either side; retain rating 3 and flag this concurrent conflict under the public policy. \\
    \addlinespace[3pt]
    \rowcolor{black!2.5}
    A2
      & Reconciliation now includes retirement and restoration, rather than only edits to active photos. An explicit source retirement may conflict with later destination edits.
      & Check archive status alongside later edits. Keep photos active when retirement conflicts with new edits, apply native archive or restore operations, and check the state after rescanning. \\
    \addlinespace[3pt]
    A3
      & File grouping is no longer fixed. The child introduces a source-side split, an existing destination-side split to preserve, and a source-side change of the primary rendition.
      & Compare the three grouping states and primary choices. Split groups or switch primary files in PhotoPrism, re-identify files by digest, and update each group's curation and sidecars. \\

    \addlinespace[5pt]
    \rowcolor{darker!7}
    \multicolumn{3}{l}{\strut\textbf{Branch B: recovering photo identity across systems}} \\
    \addlinespace[3pt]
    B1
      & Classified file relations are replaced by evidence in media metadata. True companions may have different names, while similar names or timestamps may belong to distinct captures.
      & Extract capture identifiers, Live Photo identifiers, and document lineage. Resolve logical groups and reject misleading matches before choosing a target file layout. \\
    \addlinespace[3pt]
    \rowcolor{black!2.5}
    B2
      & The destination now contains independently curated photos and target-only material. Resolving source identities no longer makes a source-authoritative overwrite safe.
      & Match photos across libraries, then compare checkpoint, source, and destination metadata. Apply source changes while preserving destination edits, independent collection changes, and target-only photos. \\
    \addlinespace[3pt]
    B3
      & Three destination representations no longer retain the exact baseline bytes. A transformed coastal PNG shares the source JPEG/RAW pair's identifier, while a similar target-only image has a different identifier.
      & Inspect identity evidence on both sides before merging curation. Link transformed representations, preserve the unrelated lookalike separately, and use native indexing results to check the grouping. \\
    \bottomrule
  \end{tabularx}
  \arrayrulecolor{black}
  \endgroup
\end{table}

\paragraph{Branch A: reconciling changes in library state.}
This branch extends reconciliation from metadata edits to retirement and restoration, then to file grouping and primary-rendition selection.
Each generation adds decisions that must precede ordinary updates.
For example, later destination edits may prevent the agent from applying a source retirement directly.
Likewise, splitting a file group in PhotoPrism can change file ownership and paths, so files must be re-identified before their curation and sidecars are assigned to the resulting groups.


\paragraph{Branch B: recovering identity across systems.}
This branch first requires the agent to infer file relationships from media evidence, then combines identity recovery with incremental synchronization into an independently edited destination.
The final generation extends identity reasoning to both systems when transformed destination files no longer match the checkpoint bytes.
The agent must establish which files represent the same photo before deciding which edits to merge.
Native indexing and rescanning then test whether these relationships and edits remain stable in the reconstructed library.


\subsection{Hotel Event Scheduling}
\label{app:evolution-hotel}

The original task asks the agent to coordinate lodging, activities, venues, catering, resources, and billing in a terminal-accessible hotel management system.
All four versions retain the same ten activities, while three successive extensions introduce attendee flows, incident recovery, and resource logistics (Table~\ref{tab:app-hotel-evolution}).


\begin{table}[!htb]
  \centering
  \setlength{\belowcaptionskip}{6pt}
  \caption{Three generations of hotel event-planning evolution, retaining the same ten activities.}
  \label{tab:app-hotel-evolution}
  \begingroup
  \normalfont
  \fontsize{9}{11}\selectfont
  \setlength{\tabcolsep}{5pt}
  \renewcommand{\arraystretch}{1.12}
  \setlength{\extrarowheight}{2pt}
  \renewcommand{\tabularxcolumn}[1]{m{#1}}
  \setlength{\heavyrulewidth}{0.7pt}
  \setlength{\lightrulewidth}{0.35pt}
  \arrayrulecolor{black!60}
  \begin{tabularx}{\linewidth}{
    >{\centering\arraybackslash\bfseries}m{32pt}
    >{\raggedright\arraybackslash}X
    >{\raggedright\arraybackslash}X}
    \toprule
    \rowcolor{nblue!6}
    \textbf{Version} & \textbf{Change from the parent} & \textbf{Additional work} \\
    \midrule

    \rowcolor{darker!7}
    \multicolumn{3}{l}{\strut\textbf{Hotel event scheduling}} \\
    \addlinespace[3pt]
    1
      & Attendee groups and directed travel times are added to the existing venue, timing, and resource constraints. The end time and location of one activity affect arrival at the next.
      & Check travel time between each group's consecutive activities and adjust venues or times when needed. Verify that each group can complete its full schedule, not just attend individual activities. \\
    \addlinespace[3pt]
    \rowcolor{black!2.5}
    2
      & Initial planning becomes recovery of a published schedule after Harbor Hall closes. Lodging and billing are completed baseline state to preserve, not work to repeat.
      & Trace the effects of venue changes and minimize disruption. Jointly revise affected assignments while preserving existing commitments, then commit a complete recovery plan atomically. \\
    \addlinespace[3pt]
    3
      & Resource quantities are refined into specific installed equipment, mobile units, and crew, with compatibility, external commitments, and strike, transfer, and installation times.
      & Check resource compatibility and calculate when equipment is ready after prior use, teardown, transfer, and installation. Revise venues, times, and resource bundles when these checks expose a conflict. \\
    \bottomrule
  \end{tabularx}
  \arrayrulecolor{black}
  \endgroup
\end{table}

\paragraph{A local disruption requires coordinated recovery.}
In the second generation, Harbor Hall's closure directly affects lunch and Workshop A.
Moving lunch to the Grand Ballroom, however, conflicts with the plenary session's occupancy, including setup and teardown buffers.
In the independent run, the agent moves the plenary to Seaview, lunch to the Ballroom, and Workshop A to Maple.
Thus, two directly affected activities lead to three venue changes, which are committed in one atomic revision without changing any activity times.


\paragraph{Resource feedback changes the recovery plan.}
In the third generation, two complete candidates are rejected because shared equipment cannot serve the proposed activity sequence.
A mobile display, for example, is released at 16:30 but needs another 15 minutes for teardown, 20 for transfer, and 20 for installation.
Its earliest ready time of 17:25 cannot support Workshop A's original 17:15 start.
The accepted plan moves the activity to Oak, combines its installed whiteboard with the mobile display, and shifts the start to 17:30 on the public 15-minute planning grid.


Across these cases, evolution changes both what the agent must determine and how those decisions depend on earlier work.
New requirements connect identity recovery to metadata merging, venue changes to attendee flows, and resource availability to schedule revision, making dependency tracing, feedback-driven revision, and integrated verification part of completing the original task.


\section{Prompt Templates}
\label{app:prompt_templates}


\subsection{Trajectory Filtering Prompt}
\label{app:trajectory_filtering_prompt}
The following prompt implements the three filtering criteria in
Appendix~\ref{app:trajectory_filtering} by reviewing complete trajectories
and returning an evidence-backed \texttt{KEEP}/\texttt{DROP} decision
for each input record.

\begin{tcolorbox}[
  paperprompt,
  title={Trajectory Filtering Prompt},
  notitle after break
]
You are an expert reviewer of terminal-agent trajectories for
supervised fine-tuning. Your task is to determine whether each complete
multi-turn trajectory provides a reliable and useful training example.

Read the entire conversation. Do not substitute excerpts, keyword
counts, command counts, or programmatic summaries for the original
trajectory. Base your judgment exclusively on the trajectory itself.
Do not access task packages, private verifiers, external benchmarks,
or the internet.

\promptheading{Understanding the trajectory}

\begin{itemize}[label={-},leftmargin=1em,labelsep=0.4em,
  topsep=0pt,itemsep=1pt,parsep=0pt,partopsep=0pt]
\item The initial user message typically contains the task description
  and the agent interaction protocol.
\item Subsequent user messages usually contain terminal observations
  produced by preceding commands, rather than new user requests.
\item Assistant messages contain reasoning, plans, and terminal actions.
  One assistant turn may contain multiple commands.
\item Reasoning, including \texttt{<think>}, analysis, and plan content, is not
  itself evidence of execution. Evaluate it together with the actual
  actions and subsequent observations.
\item A Terminus-2 trajectory may end with the terminal observation from
  its final command batch. This is a normal ending and does not require
  an additional assistant confirmation.
\item Final success or failure is not sufficient to determine training
  value. An incomplete trajectory or an honest stopping point at an
  unresolved obstacle may still provide a useful demonstration.
\end{itemize}

Evaluate only the following three dimensions.

\promptheading{1. Trajectory validity}

Is the trajectory coherent and correctly ordered? Can actions be
reliably associated with their terminal observations? Check for genuine
context truncation, severe format corruption, execution in the wrong
container, infrastructure failures, explicit refusal, or private-answer
leakage that compromises the demonstration.

Do not reject an entire trajectory because of one or two isolated
invalid JSON responses, malformed commands, or parser errors. If the
agent recognizes the problem, restores valid interaction, and continues
with a clear action\textendash observation chain, this constitutes a learnable
recovery process. Treat protocol errors as an overall validity defect
when they recur without reliable recovery or make action\textendash observation
correspondence uninterpretable.

Protocol recovery does not excuse independent integrity violations.
Reject trajectories whose core solution relies on private answers or
hidden-test leakage, internal-state tampering, deleting runtime or
database state to reset the task, or bypassing required task interfaces.

Distinguish failure to finish from lack of training value. An unfinished
trajectory can be retained when its main content demonstrates reliable
environment understanding, root-cause diagnosis, justified implementation,
feedback-based elimination of alternatives, correction of assumptions,
or identification of unresolved obstacles.

An isolated incorrect final judgment or unsupported completion claim
does not automatically invalidate all preceding work. Determine whether
the error compromises the core action chain. Retention is possible when
the main process remains reliable and useful. Reject the trajectory when
an incorrect strategy dominates the solution, central conclusions
contradict observations, or the overall example primarily demonstrates
an invalid way to complete the task.

\promptheading{2. Environment-grounded progress}

Are important actions supported by actual terminal feedback? Does the
agent use files, logs, command outputs, errors, and test results to
advance the task and adjust its approach when necessary?

Identify persistent disregard of feedback, repeated ineffective
commands, random attempts, unmotivated minor edits, and unsupported
completion claims. Evaluate these behaviors in context: unsuccessful
attempts can be valuable when their observations inform subsequent
diagnosis, revision, or recovery.

\promptheading{3. Multi-turn training value}

Does the trajectory demonstrate a multi-turn terminal problem-solving
process worth learning? Do interactions contribute meaningful evidence,
implementation, state changes, decisions, verification, or recovery?

Consider whether the apparent interaction could be trivially replaced
by one obvious edit, a fixed script, or a mechanical loop. Do not count
SSH, Docker, or Harbor infrastructure operations, waiting, repeated
polling, or irrelevant directory browsing as evidence of task difficulty.

Do not judge training value solely from trajectory length, command
count, the presence of mistakes, or final task success.

\promptheading{Decision}

\textbf{KEEP:}
The trajectory is valid, grounded in the environment, and provides
sufficient multi-turn training value. Final failure, incomplete
execution, or an honest stopping point does not preclude KEEP when
the main process is reliable and instructive.

\textbf{DROP:}
The trajectory is globally invalid; its core solution relies on
leakage, state tampering, or interface bypass; its behavior substantially
degenerates or lacks environmental grounding; or errors dominate the
main action chain with little reliable multi-turn learning value.

Do not assign DROP solely because the task was not completed.

\promptheading{Disputed cases}

If prior review claims are provided, treat them as hypotheses to verify,
not as facts or votes. Return to the original conversation to establish
whether the relevant commands were executed, whether observations
support the claims, whether the issue was subsequently repaired, and
whether it compromises the core solution. Explain how the trajectory
evidence resolves the disagreement.

\promptheading{Output}

Return exactly one review for each input record:

\begin{lstlisting}[
  style=prompt,
  basicstyle=\fontencoding{T1}\ttfamily\fontsize{8.5}{9.5}\selectfont,
  columns=fullflexible,keepspaces=true,breakatwhitespace=true,
  aboveskip=2pt,belowskip=2pt]
{
  "record_id": "<input record ID>",
  "verdict": "KEEP or DROP",
  "reason": "<concise justification>",
  "evidence": [
    "<specific evidence from the trajectory>",
    "<specific evidence from the trajectory>"
  ]
}
\end{lstlisting}

Copy \texttt{record\_id} exactly. Provide two to four concrete pieces of evidence.
Do not omit, duplicate, or invent records.
\end{tcolorbox}

\subsection{Benchmark Decontamination Prompt}
\label{app:benchmark_decontamination_prompt}
The following prompt implements the
candidate-pair assessment in Appendix~\ref{app:benchmark_decontamination}
by comparing complete task descriptions and returning an evidence-backed
\texttt{CLEAN}, \texttt{CONTAMINATED}, or \texttt{UNCERTAIN} verdict.


\begin{tcolorbox}[
  paperprompt,
  title={Benchmark Decontamination Prompt},
  notitle after break
]
\promptheading{Role and objective}

You are an expert reviewer of contamination between training data
and evaluation benchmarks. Compare each candidate Terminal/CLI task
with its paired public evaluation problem to determine whether the
candidate leaks, copies, or substantially reproduces that problem.

Read the complete descriptions of every pair in
\texttt{assigned-pairs.jsonl}. Similarity scores are retrieval signals,
not conclusions.

%
%

\promptheading{Assessment criteria}

Determine whether exposure to the training candidate would make the
evaluation problem substantially easier because the two are the same
or nearly the same task. Compare their core objectives, initial problems,
key objects or data, required transformations, distinctive constraints,
key solution paths, and final completion criteria.

\begin{itemize}[label={-},leftmargin=1em,labelsep=0.4em,
  topsep=0pt,itemsep=1pt,parsep=0pt,partopsep=0pt]
\item \textbf{CONTAMINATED:} A direct copy or minor rewording, or a renamed
  task that retains sufficiently distinctive instance details,
  constraint combinations, and solution structure to remain
  substantially the same problem.
\item \textbf{CLEAN:} The tasks share only a domain, tools, general
  terminology, or common professional work patterns; their specific
  instances, states, constraints, and required work are independent.
\item \textbf{UNCERTAIN:} The evidence is genuinely close, but the public
  descriptions alone do not permit a reliable distinction. Do not
  classify ordinary topical similarity as UNCERTAIN.
\end{itemize}

Phrase overlap arising only from templates, common interfaces,
mathematical notation, or standard industry expressions does not
constitute contamination. Conversely, the absence of long n-gram
overlap does not rule out a systematic rewrite of the same instance.

%
%
%

\promptheading{Output}

Write exactly one record for each \texttt{pair\_id} to
\texttt{reviews.jsonl}:

\begin{lstlisting}[
  style=prompt,
  basicstyle=\fontencoding{T1}\ttfamily\fontsize{8.5}{9.5}\selectfont,
  columns=fullflexible,keepspaces=true,breakatwhitespace=true,
  aboveskip=2pt,belowskip=2pt]
{
  "pair_id": "pair-...",
  "verdict": "CLEAN",
  "reason": "<core justification>",
  "evidence": [
    "<specific comparative evidence>",
    "<specific comparative evidence>"
  ]
}
\end{lstlisting}

Copy \texttt{pair\_id} exactly. The verdict must be
\texttt{CLEAN}, \texttt{CONTAMINATED}, or \texttt{UNCERTAIN}.
Write the complete output to a temporary file, then rename it atomically.
Check that the JSONL is parseable and that no records are missing or
duplicated.

Now read \texttt{assigned-pairs.jsonl} and complete the review.

%
%
%
%
\end{tcolorbox}

\end{document}